\documentclass[final,journal]{IEEEtran}
\usepackage{amsmath,graphicx}
\usepackage{tabu}
\usepackage{nccmath}
\usepackage{cleveref}
\usepackage{bbm}
\usepackage{graphbox}
\usepackage[font=small]{subfig}
\usepackage{nomencl}
\usepackage{epstopdf}
\usepackage[utf8]{inputenc}
\usepackage[style =ieee,minnames=2,maxnames=2,backend=bibtex]{biblatex}
\makenomenclature

\usepackage[belowskip=-3pt,aboveskip=4pt,font=small,labelfont=bf]{caption}
\crefrangeformat{figure}{Figs.~#3#1#4--#5#2#6}
\crefrangeformat{section}{Secs.~#3#1#4--#5#2#6}
\crefrangeformat{table}{Tabs.~#3#1#4--#5#2#6}

\DeclareMathOperator*{\argmin}{arg\,min}
\DeclareMathOperator*{\argmax}{arg\,max}

\begin{document}
\title{Robust Deep Graph Based Learning for Binary Classification}

\author{$^{1}$Minxiang~Ye, $^{1}$Vladimir~Stankovic,~\IEEEmembership{Senior Member,~IEEE,} $^{1}$Lina~Stankovic,~\IEEEmembership{Senior Member,~IEEE,}\\ $^{2}$Gene Cheung,~\IEEEmembership{Senior Member,~IEEE}

\thanks{The authors $^{1}$ are with the Department of Electronic \& Electrical Engineering, University of Strathclyde, Glasgow, G1 1XW, UK.}
\thanks{The author $^{2}$ is with the Department of Electrical Engineering \& Computer Science, York University, Toronto, M3J 1P3, Canada.}}

\maketitle

\begin{abstract}
Convolutional neural network (CNN)-based feature learning has become state of the art, since given sufficient training data, CNN can significantly outperform traditional methods for various classification tasks. 
However, feature learning becomes more difficult if some training labels are noisy. 
With traditional regularization techniques, CNN often overfits to the noisy training labels, resulting in sub-par classification performance. 
In this paper, we propose a robust binary classifier, based on CNNs, to learn deep metric functions, which are then used to construct an optimal underlying graph structure used to clean noisy labels via graph Laplacian regularization (GLR). 
GLR is posed as a convex maximum a posteriori (MAP) problem solved via convex quadratic programming (QP). 
To penalize samples around the decision boundary, we propose two regularized loss functions for semi-supervised learning. 
The binary classification experiments on three datasets, varying in number and type of features, demonstrate that given a noisy training dataset, our proposed networks outperform several state-of-the-art classifiers, including label-noise robust support vector machine, CNNs with three different robust loss functions, model-based GLR, and dynamic graph CNN classifiers.
	\end{abstract}
	\begin{IEEEkeywords}
		deep learning, graph Laplacian regularization, binary classification, semi-supervised learning
	\end{IEEEkeywords}
	\IEEEpeerreviewmaketitle
	
	\nomenclature[01]{$\mathbf{X}$}{A set of observations}
	\nomenclature[02]{$r$}{GLR iteration number}
	\nomenclature[03]{$\mathbf{Y}^r$}{Labels corresponding to $\mathbf{X}$ at $r$-th GLR iteration}
	\nomenclature[04]{$\mathbf{\dot{X}},\mathbf{\dot{Y}}^r$}{A subset of $\mathbf{Y}^r$ corresponding to the training samples in $\mathbf{\dot{X}}$}
	\nomenclature[05]{$\mathbf{E}=\{e_{i,j}\}$}{Binary matrix that represents the edge connectivity with each entry $e_{i,j}$ corresponding the edge connecting node $i$ to node $j$}
	\nomenclature[06]{$\mathbf{W}=\{w_{i,j}\}$}{A weight matrix with each entry $w_{i,j}$ assigned to the corresponding edge $e_{i,j}$}
	\nomenclature[07]{$\mathbf{G}=(\mathbf{\Psi}, \mathbf{E},\mathbf{W})$}{A undirected graph that comprises a set of nodes $\mathbf{\Psi}$, edge matrix $\mathbf{E}$ and corresponding weights $\mathbf{W}$}
	\nomenclature[08]{$\mathbf{L}$}{Combinatorial graph Laplacian matrix}
	\nomenclature[09]{$\mathbf{A}$}{Adjacency matrix}
	\nomenclature[10]{$\mathbf{D}$}{Degree matrix}
	\nomenclature[11]{$d_{max}$}{Maximum degree of node in $\mathbf{G}$}
	\nomenclature[12]{$\mathcal{D}(\cdot), \mathcal{H}_{\mathbf{U}}(\cdot)$}{Deep feature maps}
	\nomenclature[13]{$\mathcal{V}^r(\cdot),\mathcal{C}^r(\cdot)$}{Deep feature maps associated with $r$-th GLR iteration}
	\nomenclature[14]{$\mathcal{Z}_{D}(\cdot), \mathcal{Z}_{\mathcal{H}_{\mathbf{U}}}(\cdot)$}{Shallow feature maps in $\mathcal{D}(\cdot)$ and $\mathcal{H}_{\mathbf{U}}(\cdot)$}
	\nomenclature[15]{$f^r(x)$}{Observations associated with $r$-th GLR iteration}
	\nomenclature[16]{$g(x)$}{Observations associated with graph update}
	\nomenclature[17]{$x_a$}{A random node $a$ selected from $\mathbf{X}$}
	\nomenclature[18]{$x_p$}{A random node $p$ selected from $\mathbf{X}$, with same label as $x_a$}
	\nomenclature[19]{$x_n$}{A random node $n$ selected from $\mathbf{X}$, with opposite label as $x_a$}
	\nomenclature[20]{$\mathbf{P}$}{A set of edges linked between nodes with same label}
	\nomenclature[21]{$\mathbf{Q}$}{A set of edges linked between nodes with opposite label}
	\nomenclature[22]{$\alpha_{\mathbf{E}}, \alpha_{\mathbf{W}}$}{Minimum margin between deep metric based distances of $\mathbf{P}$ and $\mathbf{Q}$ edges for $Loss_{\mathbf{E}}$ and $Loss_{\mathbf{W}}$}
	\nomenclature[23]{${\mathbf{\gamma}}^r$}{Maximum number of nodes connected to each node in graph $\mathbf{G}^r$}
	\nomenclature[24]{$\mathbf{y}_i^1$}{Encoded vector corresponding to the label $y_i$ for node $i$}
	\nomenclature[25]{$\mathbf{y}_{\mathcal{U}}^1$}{Encoded matrix corresponding to the labels for neighboring nodes}
	\nomenclature[26]{$\Theta$}{Activation function that estimates how much attention is paid on each edge}
	\nomenclature[27]{$\varPi=\{\pi_{i,j}\}$}{Attention matrix with each entry $\pi_{i,j}$ corresponds to edge loss}
	\nomenclature[28]{$\pi_{a,p}, \pi_{a,n}$}{Attentions on $\mathbf{P}$ and $\mathbf{Q}$ edges}
	\nomenclature[29]{$\Phi$}{Edge Attention Activation}
	\nomenclature[30]{$\kappa$}{Conditional number}
	\nomenclature[31]{${\mu}^r$}{Smoothness prior factor}
	\nomenclature[32]{$\beta$}{Soft edge connectivity for graph update}
	\nomenclature[33]{${\varepsilon}^r$}{Thresholds in $\Theta$ and $\Phi$}
	\printnomenclature[2.5cm]

	\section{Introduction}
	\label{sec:intro}
    Supervised and semi-supervised deep learning techniques have shown excellent performance for feature extraction and classification tasks \cite{krizhevsky2012_imagenet}, but are particularly sensitive to the quality of the training dataset, since they tend to overfit the models when learning from incorrect labels \cite{sukhbaatar2014_noisecnn, patrini2017_losscorrection,ma2018_d2l}. Since labels assigned to training samples are sometimes corrupted, studies on classifier learning with ``noisy" labels are of practical importance \cite{tang2011_knnnoise, frenay2014_noisesurvey, wang2018_iterative}.
	
	Conventional approaches to overcome model overfitting, based on various regularization techniques, e.g., $l_1-$ or $l_2$-norm penalty on weights \cite{lemberger2017_generalization}, dropout \cite{hinton2012_dropout}, batch normalization \cite{ioffe2015_batchnorm}, skip-connections \cite{he2016_renet,huang2017_densenet} etc., are not effective in mitigating the effects of incorrect labels. 
	This has given rise to different approaches to learning using noisy training labels. 
	These approaches can be grouped into methods based on: (a) inserting additional ``trusted" labels, e.g., \cite{xiao2015_massivenoise}, \cite{hendrycks2018_trust}, and (b) loss function correction, e.g., \cite{masnadi2009_savage}, \cite{reed2014_bootstrap}, \cite{goldberger2016_noiseadaptation}. In 	\cite{xiao2015_massivenoise}, a probabilistic model is integrated into the deep neural network (DNN) to correct noisy labels. Similarly, in \cite{hendrycks2018_trust}, a loss correction technique is introduced to mitigate the unreliability of noisy training labels. However, these methods require clean data to prevent the models from drifting away.  A robust loss function that is less sensitive to label outliers is introduced in \cite{masnadi2009_savage}. A combination of training labels and predicted labels is used in \cite{reed2014_bootstrap} to avoid directly modeling the noisy training labels, but requires pre-training to achieve good results. \cite{goldberger2016_noiseadaptation} adds another softmax layer to further augment the correction model via a noise transition matrix, which is hard to estimate in practice, especially in multi-class classification problems.
	
	An alternative approach is to restore corrupted training labels by representing them as piece-wise smooth signals on graphs and applying a graph signal smoothness prior \cite{Sandryhaila2013_gr,gong2015_deformedlap,gene2018_negglrgu,cheng2018_altergraphclassifier}. In \cite{jin2018_dglr}, an image denoising scheme is proposed using graph Laplacian regularization (GLR), given either small- or large-scale datasets. 
	Building on \cite{jin2018_dglr},
	 \cite{jiang2018_glrcnn} integrates GLR into the DNN to perform semi-supervised classification of nodes in a citation network, considering, during label propagation, the local consistency of nodes with similar features. 
	 However, these hybrid methods \cite{jiang2018_glrcnn}, \cite{jin2018_dglr}, are only performed and evaluated for a fixed graph (i.e., a 2D grid of image pixels). 
	 Without prior knowledge of the graph structure, more recently, \cite{ye2019_dglr} proposes a regularized triplet loss correction function to mitigate the effects of insufficient clean training samples via a sparse $K$-nearest neighbor (KNN) graph construction and GLR.
	
	In this paper, we further extend the previous studies on binary classification in the presence of noisy labels with DNN-based classifier learning using GLR without prior knowledge of the underlying graph. 
	We propose an end-to-end trainable network for semi-supervised binary classification, that incorporates an attention mechanism to capture important feature information and guide learning through a regularized triplet loss and iterative GLR, given a sufficient number of training samples, many of them corrupted. 
	In summary, the main contributions of this paper are:
	\begin{enumerate}
	\item To avoid over-fitting the classifier, we propose a graph-based regularized loss function that incorporates attention mechanism to regularize the proposed network;
	\item To mitigate the effects of noisy training labels and improve the reliability of classifier learning, we develop a graph-based semi-supervised classifier that iteratively performs both online denoising of training labels and classification at the same time;
	\item To assign a degree of freedom for graph connectivity learning that is robust to noisy labels, we introduce a graph update procedure to better reflect the node-to-node correlation based on convolution of iteratively updated edges.
	\item A complete semi-supervised binary classification scheme, tested for a range of classification tasks and benchmarked against state-of-the-art classifiers in the presence of noisy training labels.
	\end{enumerate}
	
	Our proposed network is evaluated against several classic and state-of-art methods, designed specifically for the ``noisy label'' problem, These methods include support vector machines (SVM) \cite{cortes1995_svm}, convolutional neural network (CNN), dynamic graph CNN \cite{wang2018_dyngcnn}, deep metric-based KNN classifier \cite{elad2014_triplet}, label noise robust SVM \cite{biggio2011_lnrsvm}, GLR-based approach \cite{gene2018_negglrgu} and CNN with robust loss corrections \cite{masnadi2009_savage, reed2014_bootstrap, ma2018_d2l}. For the ablation study of the overall network, we disable different proposed components to study their effect on performance.
	
	The rest of the paper is structured as follows. First, an overview of related work is provided in Section~ \ref{sec:prior}. Then, in Section \ref{sec:classifierlearning}, we introduce the notation and formulate the ``noisy label'' classifier learning problem. In Section \ref{sec:dynGLR}, we describe the implementation of the proposed network. Finally, in Section~\ref{sec:simulations} we evaluate and discuss the performance of the proposed classifier against state-of-the-art methods and present findings of the ablation study for three different datasets.
	
	\section{Related Work}
	\label{sec:prior}
	In this section, we first provide an overview of the related work on robust graph-based classifier learning and robust DNN-based classifier learning, in the presence of ``noisy labels''. Then we discuss graph-based methods integrated with DNN that do not consider noisy data. Finally, the weaknesses of the state-of-the-art classifiers are discussed to motivate the present work.
	
	\subsection{Robust Graph-based Learning}
	 A label propagation method is proposed in \cite{speriosu2011_glp} to evenly spread, throughout the graph, label distributions from selected labeled nodes, which are usually noisy and with heuristic information. A KNN-sparse graph-based semi-supervised learning approach is proposed in \cite{tang2011_knnglp} to remove most of the semantically-unrelated edges and adopt a refinement strategy to handle noisy labels. 
	
	To achieve more robust binary classification, in \cite{gene2018_negglrgu}, negative edge weights are introduced into the graph to separate the nodes in two different clusters. A perturbation matrix is found to perform generalized GLR for binary classification via iterative re-weighted least squares strategy \cite{daubechies2010_irls}. The results demonstrate the applicability of negative edge weights for robust graph-based classifier learning for small amount of data without learning feature representation. We evaluate this approach in this paper when sufficient but noisy training labels are provided.
	
	\subsection{Robust DNN-based Classifier Learning}
	Many studies investigate methods to accommodate a wide range of label noise levels and types, which often focus on data augmentation, network design-based regularization and loss correction. Data augmentation techniques have been successfully used in \cite{prest2012_da, misra2015_da, kuznetsova2015_da} to automatically annotate unlabeled samples and use these samples for retraining.
	In \cite{ren2018_robustreweight}, training examples are assigned weights by a proposed meta-learning algorithm to minimize the loss on a clean unbiased validation set based on gradient direction. 
	
	The effectiveness of dropout regularization for cleaning noisy labels is shown in \cite{jindal2016_dropout}. For image classification, \cite{rolnick2017_robust_massivenoise} indicate that increasing the batch size and downscaling the learning rate is a practical approach to mitigate the effects of label noise for a DNN-based classifier, given a sufficiently large training set. 
	
	 The loss correction approach of  \cite{masnadi2009_savage} proposes a boosting algorithm `SavageBoost' that is less sensitive to outliers and converges faster than conventional methods, such as Ada, Real, or LogitBoost. A noise-aware model is formulated in \cite{mnih2012_robust} to handle label omission and registration errors for improving labeling of aerial images. A dimensionality-driven learning strategy is discussed in \cite{ma2018_d2l} to avoid overfitting by identifying the transition from an early learning stage of dimensionality compression to an overfitting learning stage when the local intrinsic dimensionality steadily increases. Unlike the above loss correction studies to handle noisy and incomplete labeling, \cite{reed2014_bootstrap} use a combination of training labels and the prediction from the current model to update the training targets and perform weakly-supervised learning.
	Similarly, \cite{xiao2015_noisylabel} integrate the Expectation-Maximization (EM) algorithm into CNN to detect and correct noisy labels, but require a properly pre-trained model. \cite{wang2018_iterative} propose an iterative learning framework to facilitate `robustness to label noise' classifier learning by jointly performing iterative label detection, discriminative feature learning and re-weighting.

    \subsection{Graph-based classifier learning with DNN}
	Recent years have seen integration of graph-based learning with deep learning. 
	Given a fixed graph structure, \cite{joan2013_sgcnn, kipf2016_semigcnn, defferrard2016_gcnn} design CNNs for feature learning by feeding a polynomial of the graph Laplacian. 
	\cite{zhang2018_edgeconv} adopt edge convolution to learn combinational spatial features from neighboring nodes given a fixed skeleton graph. Based on the ideas of edge convolution, \cite{wang2018_dyngcnn} propose a deeper CNN model to learn the underlying KNN graph structure of point cloud data by iteratively updating the graph. The results demonstrate the capability of edge convolution for feature generalisation on point-cloud data. \cite{jin2018_dglr} propose a deep image denoising framework that couples encapsulation of the fully-differentiable Graph Laplacian regularization layer and learning 8-connected pixel adjacency graph structures via CNNs. The results indicate that given a small dataset, the method of \cite{jin2018_dglr} outperforms state-of-the-art approaches. 
	
	The problem of insufficient data or incorrect training labels has not been investigated in the above graph-based hybrid methods for classifier learning. For incomplete or imprecise categories of tags (observations) in the training samples, \cite{mojoo2017_dglrcnn} combine CNN and GLR using the sum of the cross-entropy loss and the GLR term for multi-label image annotation, where CNN is used to construct the fully-connected similarity-based graph. Unlike \cite{mojoo2017_dglrcnn}, in our conference paper \cite{ye2019_dglr}, we integrate GLR into CNN with a graph-based loss correction function to tackle the problem of insufficient training samples through semi-supervised graph learning.
	
	\subsection{Novelty with respect to reviewed literature}
	In this paper, we further extend our conference contribution \cite{ye2019_dglr} to a more generalized end-to-end CNN-based approach given noisy binary classifier signal, to perform iteratively GLR (similar to \cite{jin2018_dglr}) as a classifier signal restoration operator, update the underlying graph and regularize CNNs. Compared to the previous graph-based classifiers \cite{speriosu2011_glp,Sandryhaila2013_gr,joan2013_sgcnn,ekambaram2013_wrssl,gadde2014_assl,kipf2016_semigcnn,gene2018_negglrgu,jiang2018_glrcnn,cheng2018_altergraphclassifier}, by adopting edge convolution, iteratively updating graph and operating GLR, we learn a deeper feature representation, and assign the degree of freedom for learning the underlying data structure. Given noisy training labels, in contrast to the classical robust DNN-based classifiers \cite{masnadi2009_savage,reed2014_bootstrap,xiao2015_noisylabel,wang2018_iterative,ren2018_robustreweight,ma2018_d2l}, we bring together the regularization benefits of GLR and the benefits of the proposed loss functions to perform more robust deep metric learning. We further adopt a rank-sampling strategy to find those training samples with high predictive performance that benefits inference.
	
	\section{Robust Deep Graph Based Classifier Learning}
	\label{sec:classifierlearning}
	In this section, we first introduce notation and formulate the robust classifier learning problem following the related work \cite{Sandryhaila2013_gr, gene2018_negglrgu, cheng2018_altergraphclassifier, jin2018_dglr, elad2014_triplet}. Then, 
	we describe the main concept behind the proposed Dynamic Graph Laplacian Regularization (DynGLR) neural network that learns robust deep feature map to effectively perform GLR when parts of the labeled data available to train the model are noisy. 
	
	\subsection{Problem Formulation and Notation}
	\label{sec:problem}
	Given observations  
	$\mathbf{{X}}=\{x_1,\ldots,x_{N}\}$, where $x_i \in \mathcal{R}^n$, $i=1,\ldots N$,
	the task of a binary classifier is to learn an approximate mapping function that maps each observation $x \in \mathbf{X}$ into a corresponding binary discrete variable $y \in \mathbf{Y}=\{y_1,\ldots,y_N\}$, called classification label, where $y_i \in \{-1, +1\}$, $i=1,\ldots, N$.
	
	Let $\mathbf{\dot{Y}}^0=\{-1,1\}^{M}=\{y_1,\ldots,y_M\} \subset \mathbf{Y}$, $0<M<N$, be a set of known (possibly noisy) labels that correspond to instances $\mathbf{\dot{X}}=\{x_1,\ldots, x_{M}\} \subset \mathbf{X}$ used for training.
	Let $\mathbf{Y}^0=\{\mathbf{\dot{Y}}^0, \mathbf{0}^{N-M}\}$, where we set to zero all $N-M$ unknown labels (to be estimated during testing).
	 
	Given $\mathbf{{X}}$, the problem addressed in this paper, is to learn the robust mapping function to assign a classification label to each observation $x \in \mathbf{X}$ when some classification labels $y \in \mathbf{\dot{Y}}^0$, used for training the model, are incorrect.

	Let $\mathbf{G}=(\mathbf{\Psi}, \mathbf{E},\mathbf{W})$ be an undirected graph, where $\mathbf{\Psi}=\{\psi_1,\ldots,\psi_N\}$ is a set of nodes, each corresponding to one instance in $\mathbf{X}$, $\mathbf{E}=\{e_{i,j}\}, i,j \in \{1,\ldots,N\}$, is a matrix representing the edge connectivity of $\mathbf{G}$; that is, $e_{i,j}=1$ if there is an edge connecting vertices $i$ and $j$ and $e_{i,j}=0$ otherwise; and each entry $w_{i,j}$ in the weight matrix $\mathbf{W}=\{w_{i,j}\}, i,j \in \{1,\ldots,N\}$ corresponds to the weight associated with edge $e_{i,j}$. Then,
	$\mathbf{Y}^0$ can be seen as a graph signal that indexes the graph $\mathbf{G}$. The combinatorial graph Laplacian matrix is given by $\mathbf{L}=\mathbf{D}-\mathbf{A}$, where $\mathbf{A}$ is a symmetric $N \times N$ adjacency matrix with each entry $a_{i,j}=max({w}_{i,j}\cdot e_{i,j}, {w}_{j,i}\cdot e_{j,i})$, and $\mathbf{D}$ is a degree matrix with entries $d_{i,i}=\sum_{j=1}^{N} a_{i,j}$, and $d_{i,j}=0$ for $i \ne j$.
	
    Similarly to \cite{elad2014_triplet}, we define {\it triplets} as observations $(x_a, x_p, x_n)$, $x_a, x_p, x_n \in \mathbf{X}$ corresponding to vertices $\psi_a, \psi_p, \psi_n \in \mathbf{\Psi}$, respectively, such that $y_a=y_p \ne y_n$, and $y_a,y_p,y_n \in \mathbf{\dot{Y}}$. Let $\mathbf{P}$ be a set of all edges $e_{a,p}$, such that $y_a=y_p$, and $\mathbf{Q}$ a set of all edges $e_{a,n}$, for which $y_a \ne y_n$, that is, $\mathbf{P}$ and $\mathbf{Q}$ are sets of all edges that connect nodes with the same and opposite labels, respectively.
    

	Motivated by CNNs ability to extract discriminative features and GLRs to `clean' unreliable labels, we formulate graph-based classifier learning as a two-stage learning process: (1) {\it graph learning} - extract deep feature maps, i.e., find a deep metric function that returns the most discriminative feature maps, and then generate an initial graph by learning the underlying $\mathbf{E}$ to maximize/minimize similarity between any two nodes in $\mathbf{G}$ that are indexed by the same/opposite labels. (2) {\it classifier learning} - iteratively refine the graph and effectively performing GLR to restore the corrupted classifier signal. In the following, we describe these two stages.
	
	\subsection{Initialization}
	\label{sec:representationlearning}
	 Given the observation sets $\mathbf{X}$ and corresponding, potentially noisy labels, $\mathbf{\dot{Y}^0}$, the first task is to learn a discriminative feature map $\mathcal{V}^0(\cdot)$ and generate an initial underlying graph for the learnt feature map. 
	
	Let
	$$d_{i,j}(\mathcal{V}^0)=\lVert \mathcal{V}^0(x_i)-\mathcal{V}^0(x_j)\rVert_2^2,$$ be the Euclidean distance between the corresponding feature maps.
	For a node $\psi_i \in \mathbf{\Psi}$, let $\mathcal{E}_i$ be a set containing all vertices except $\psi_i$ in ascending order with respect to the metric $d_{i,k}(\mathcal{V}^0)$, $k=1,\ldots,N-1, k\ne i$.
	Let $\mathcal{S}_i$ be a subset of $\mathcal{E}_i$ containing the first $\gamma_i$ elements of $\mathcal{E}_i$, that is, the set $\mathcal{S}_i$ contains $\gamma_i$ most correlated vertices to vertex $\psi_i$ according to metric $d_{i,k}(\mathcal{V}^0)$.
	
\begin{figure}[!htbp]
	 	\centering
	 	\includegraphics[width=0.68\linewidth]{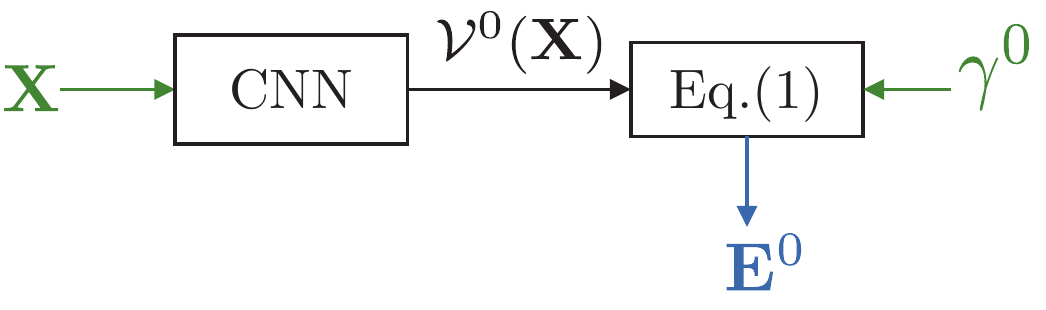}
	 	\caption{The block diagram of the unweighted graph generation scheme. $\mathcal{V}^0(\cdot)$ is a CNN-based feature map learnt by minimizing a loss function in order to reflect the node-to-node correlation. The implementation of the proposed CNN and loss function is described in Sec.~\ref{sec:gnet}.}
	 	\label{fig:generator}
	 \end{figure}

	
	To effectively perform GLR, as in \cite{gene2018_negglrgu, jin2018_dglr}, the underlying graph should be a sparsely connected graph. To control the sparsity of the resulting graph whilst maintaining connectivity, we use an indicator operator to minimize the number of $\mathbf{Q}$ edges. A typical option is a KNN indicator that keeps only a maximum of $\gamma_i$ edges for each individual node $i$, and sets others to zero.
	That is, each graph edge $e_{i,j}$ is set to:
	\begin{equation}
	\begin{aligned}
	e_{i,j}=&\begin{cases}
	1, &\text{if } {\psi}_i \in \mathcal{S}_j \text{ or } {\psi}_j \in \mathcal{S}_i \\
	0, & otherwise.
	\end{cases}
	\end{aligned}
	\label{eq:connectivity}
	\end{equation}
	 Once an optimal edge matrix $\mathbf{E}^0=\{e_{i,j}^0\}$ is computed through $\mathcal{V}^0(\cdot)$ and (\ref{eq:connectivity}), we obtain an initial undirected and unweighted graph $\mathbf{G}^0=(\mathbf{\Psi}, \mathbf{E}^0, \mathbf{W}^0=1)$. The block diagram is shown in Fig.~\ref{fig:generator}. Note that, in our implementation, we start with $\gamma_1=\cdots=\gamma_N=\gamma^0$, that is learnt as explained in Sec.~\ref{sec:gnet}.
	
	\subsection{Proposed Classifier Learning with Iterative Graph Update}
	\label{sec:glrlearning}
	If the noisy training labels are seen as a piece-wise smooth graph signal, $\mathbf{Y}^0$, then one can iteratively perform GLR for denoising the labels and performing semi-supervised classification, while refining the set of deep feature maps and the underlying graph.
	
	Let $r>0$ be the iteration index, initialized to 1, and let $\mathbf{G}^r=(\mathbf{\Psi}, \mathbf{E}^{r-1},\mathbf{W}^r)$ be the graph, with $\mathbf{W}^r$ to be learnt, and
	$\mathbf{Y}^r$ the noisy labels in the $r$-th iteration. Thus, $\mathbf{G}^1$ is an $N$-node graph with edges set by (\ref{eq:connectivity}). In the $r$-th iteration, each vertex $\psi_i$ is indexed by a label $y_i^{r-1} \in \mathbf{Y}^{r-1}$ (graph signal), and is associated to a feature vector $\mathcal{V}^r(x_i)$.
	
	 Typically, the edge weight is computed using a Gaussian kernel function with a fixed scaling factor $\sigma$, i.e., 
	$\exp\Big(-\frac{\lVert x_i-x_j \rVert_2^2}{2{\sigma}^2}\Big)$, to quantify the node-to-node correlation. Instead of using a fixed $\sigma$ as in \cite{Sandryhaila2013_gr,gene2018_negglrgu,cheng2018_altergraphclassifier}, motivated by \cite{Zelnik2004_lsk}, we introduce an auto-sigma Gaussian kernel function to assign edge weight $w_{i,j}^r$ in $\mathbf{G}^r$ by maximizing the margin between the edge weights assigned to $\mathbf{P}$-edges and $\mathbf{Q}$-edges, as:
	\begin{equation}
	\label{eq:edgeweight}
	\begin{aligned}
	\sigma^*=&\argmax_{\sigma}\Big[\exp{\Big(-\frac{{\omega}_{\{\psi_a,\psi_p\}}^2}{2\sigma^2}\Big)}
	-\exp{\Big(-\frac{{\omega}_{\{\psi_a,\psi_n\}}^2}{2\sigma^2}\Big)}\Big]\,\\
	w_{i,j}^r=&\exp\Big(-\frac{\lVert \mathcal{V}^r(x_i)-\mathcal{V}^r(x_j) \rVert_2^2}{2{\sigma^*}^2}\Big)
	\end{aligned}
	\end{equation}
	where ${\omega}_{\{\psi_a,\psi_p\}}$ and ${\omega}_{\{\psi_a,\psi_n\}}$ compute the mean Euclidean distances between nodes connected by $\mathbf{P}$-edges and $\mathbf{Q}$-edges, respectively.
	By setting the first derivative to zero, we obtain the resulting optimal $\sigma^*=\sqrt{\frac{{\omega}_{\{\psi_a,\psi_n\}}^2-{\omega}_{\{\psi_a,\psi_p\}}^2}{2\log({\omega}_{\{\psi_a,\psi_n\}}^2/{\omega}_{\{\psi_a,\psi_p\}}^2)}}$, which is used to assign edge weights of the graph.

\begin{figure}[!tbh]
		\centering
		\subfloat[The block diagram of the proposed classifier scheme. CNN is learnt by minimizing the loss function to better reflect the node-to-node correlation. The edge matrix $\mathbf{E}^{r-1}$ is used as a mask when assigning edge weights to construct adjacency matrix $\mathbf{A}^r$. We perform GLR to restore the corrupted classifier signal $\mathbf{Y}^{r-1}$ given the resulting sparse graph Laplacian $\mathbf{L}^r$ and apply the constrained smoothness prior factor ${\mu}^r$ to ensure the numerical stability of QP solver. The implementation of $\mathcal{V}^r(\cdot)$ varies depending on the data scale and the dimension of the input observations. The output is the new set of `denoised' labels $\mathbf{Y}^r$.]{\includegraphics[width=0.8\linewidth]{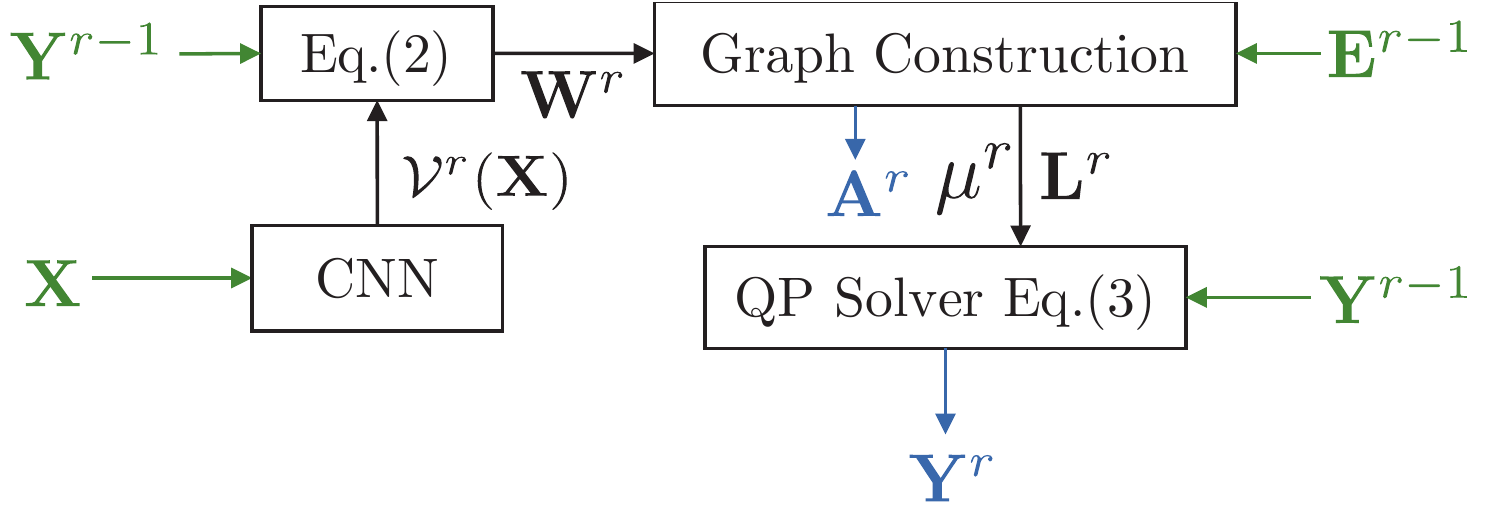}}\\
		\subfloat[The block diagram of the proposed graph update scheme. Based on the adjacency matrix $\mathbf{A}^r$  and restored classifier signal $\mathbf{Y}^r$, we learn a CNN to better refine the graph structure. The edge matrix $\mathbf{E}^r$ is updated via (\ref{eq:graphupdate}) and (\ref{eq:connectivity}) based on both the previous restored classifier signal and the regularized deep feature map. The output of this block is the new edge matrix $\mathbf{E}^r$ that will be used in the next iteration. ]{\includegraphics[width=0.8\linewidth]{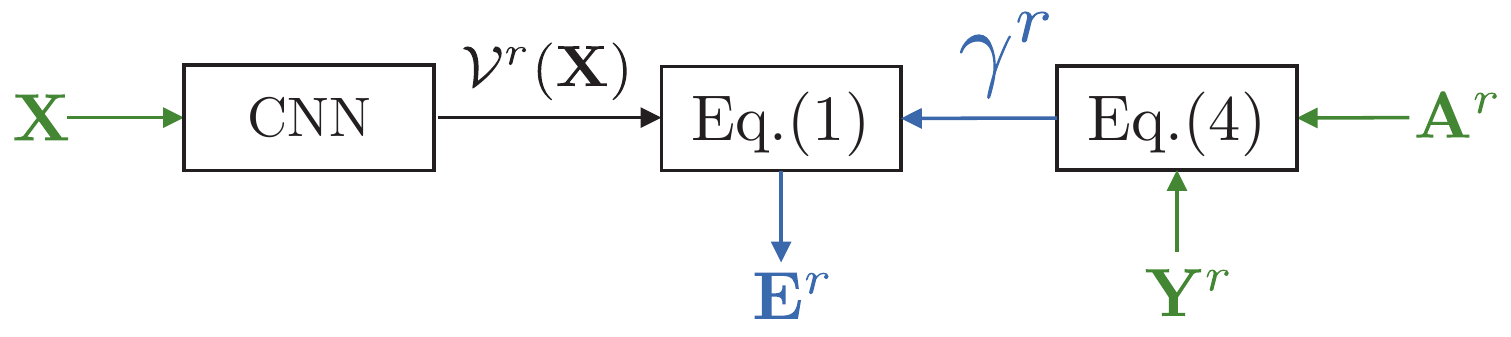}}
		\caption{The proposed graph-based classifier and graph update scheme. The green and blue colors denote input and output, respectively. The implementation details are given in Sec.~\ref{sec:wnet}.}
		\label{fig:graphupdate}
	\end{figure}

	Closely following related work \cite{gene2018_negglrgu, jin2018_dglr}, we obtain the restored classifier signal by finding the smoothest graph signal $\mathbf{Y}^{r}$ as:
	\begin{equation}
	\label{eq:glr}
	\begin{aligned}
	\mathbf{Y}^{r}=\argmin_{\mathbf{B}}~(\lVert \mathbf{Y}^{r-1}-\mathbf{B} \rVert_2^2+\mu^r \mathbf{B} \mathbf{L}^r{\mathbf{B}}^T).
	\end{aligned}
	\end{equation}
	The minimization above finds a solution that is close to the observed set of labels in the previous iteration, $\mathbf{Y}^{r-1}$, while preserving piece-wise smoothness. To guarantee that the solution $\mathbf{Y}^{r}$ to the quadratic programming (QP) problem (\ref{eq:glr}) is numerically stable, we adopt Theorem 1 from \cite{jin2018_dglr} by setting an appropriate conditional number $\kappa$. The maximum value of the smoothness prior factor ${\mu}^r$ is then calculated as: ${\mu}^r_{max}=(\kappa-1)/(2d_{max}^r)$, where $d_{max}^r$ is the maximum degree of the vertices in graph $\mathbf{G}^r$.
	 See Fig.~\ref{fig:graphupdate}(a).
	
	Between each two GLR iterations, we use CNN to refine the feature map based on the denoised label signal, $\mathbf{Y}^{r-1}$ obtained in the previous GLR iteration. See an illustration in Fig.~\ref{fig:graphupdate}(b) for the graph update after $r$-th GLR iteration. We update the individual degree of Vertex $i$ as:
	\begin{equation}
		\begin{aligned}
		\mathring{e}_{i,j}^r=&\begin{cases}
				1, &\text{if } e_{i,j}^r \in \mathbf{P}^r ~\&~a_{i,j}^r>\beta \\
				0, &\text{if } e_{i,j}^r \in \mathbf{Q}^r ~\&~a_{i,j}^r\leq \beta, \end{cases}\\
		\gamma_i^r=&\sum_{j=1}^{N} \mathring{e}_{i,j}^r,
		\end{aligned}
		\label{eq:graphupdate}
	\end{equation}
	where $\mathbf{P}^r$ and $\mathbf{Q}^r$ sets are formed based on the denoised classifier signal $\mathbf{Y}^r$. The edge $e_{i,j}^r$ is removed if it connects vertices with opposite labels or the corresponding entry to adjacency matrix is less than $\beta$, which is heuristically set to 0.1.

	
	\section{Proposed Network}
	\label{sec:dynGLR}
	Based on the concepts described in the previous section, in this section we present the algorithmic flow and describe the architecture used to implement the proposed DynGLR network.
	
	\begin{figure}[!tbh]
		\centering
		\includegraphics[width=0.88\linewidth]{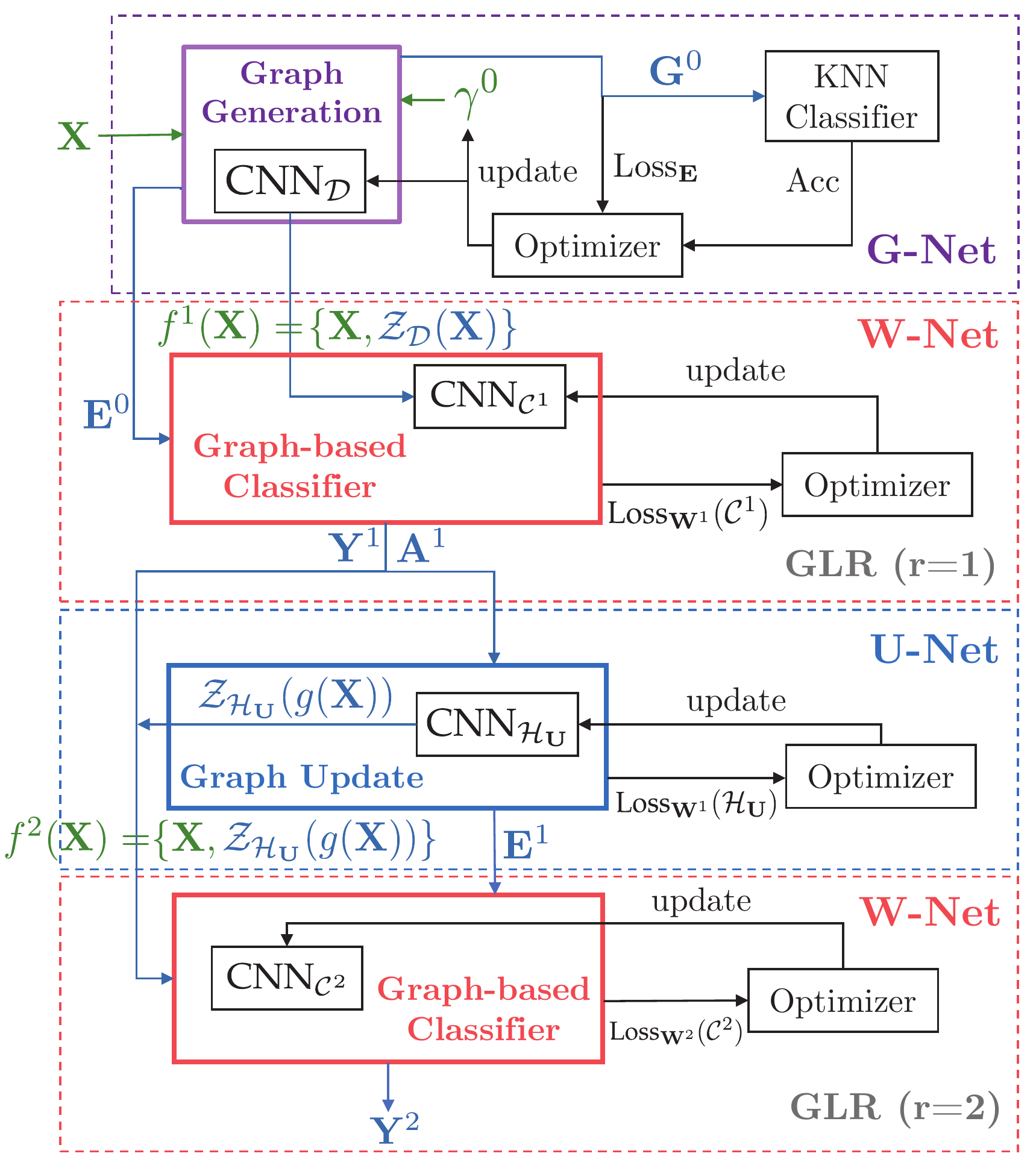}
		\caption{The overall block diagram of the proposed DynGLR-Net for $r=2$. Given observations ${\mathbf{X}}$, G-Net (see Subsec.~\ref{sec:gnet}) first learns an initial undirected and unweighted KNN-graph by minimizing $\text{Loss}_{\mathbf{E}}$. The resulting edge matrix $\mathbf{E}^0$ is then used in the following, first, GLR iteration. The learnt shallow feature map $f^1(X)=\{X,\mathcal{Z}_{\mathcal{D}}(X)\}$ is then used as input to learn a $\text{CNN}_{\mathcal{C}^1}$ network for assigning weights to the initial graph edges. Given a subset of, potentially noisy labels, ${\mathbf{\dot Y}}$, we perform GLR on the constructed undirected and weighted graph to restore the labels. The resulting restored labels are used in the following GLR iterations (see Subsec.~\ref{sec:wnet}). To assign the degree of freedom for refining graph connectivity, we update the graph edge sets by minimizing $\text{Loss}_{\mathbf{W}^1}({\mathcal{H}}_{\mathbf{U}})$ given neighbor information for each node based on the resulting denoised classifier signal from the first GLR iteration. We then reassign edge weights to the updated graph edge sets to perform better node classification in the second GLR iteration (see Subsec.~\ref{sec:unet}).}
		\label{fig:dynglr}
	\end{figure}
	The block diagram of the proposed DynGLR-Net is presented in Fig.~\ref{fig:dynglr}. Our overall network consists of three sub-networks: (1) {\bf G-Net} (graph generator network) used to learn a deep metric function to construct an undirected and unweighted KNN graph $\mathbf{G}^0=(\mathbf{\Psi},\mathbf{E}^0, \mathbf{W}^0=\mathbf{1})$. 
	(2) {\bf W-Net} (graph weighting and classifier network) used to assign edge weights $\mathbf{W}^r$ for effectively performing GLR to restore the corrupted classifier signal $\mathbf{Y}^{r}$. (3) {\bf U-Net} (graph update network) used to refine $\mathbf{E}^r$ to better reflect the node-to-node correlation based on the restored classifier signal in the previous iteration $\mathbf{Y}^{r-1}$. We clarify each network in the following subsections. 
	
	\subsection{G-Net}
	\label{sec:gnet}
\begin{figure}[!htbp]
		\centering
		\includegraphics[width=0.85\linewidth]{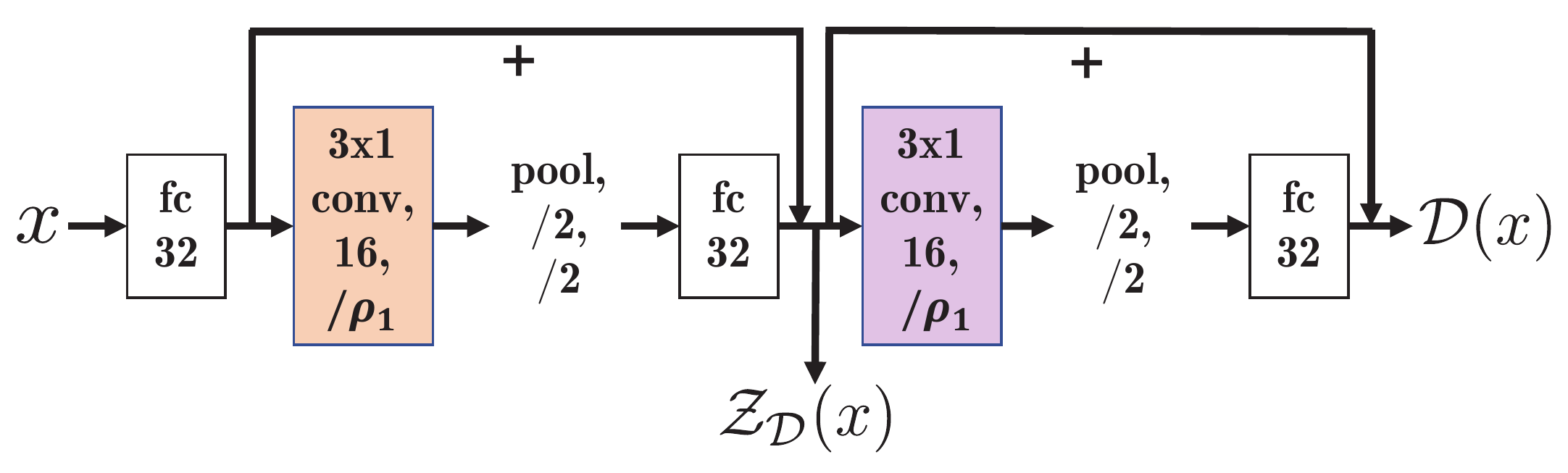}
		\caption{$\text{CNN}_{\mathcal{D}}$ neural network: `pool/q/w' refers to a max-pooling layer with q=pool size and w=stride size. `x conv y/$\rho_1$' refers to a 2D convolutional layer with y filters each with kernel size x and stride size $\rho_1$. `fc x' means the fully connected layer with x=number of neurons. The stride size $\rho_1$ varies depending on the input data (see details in Sec.~\ref{subsec:setup}).}
		\label{fig:cnnD}
	\end{figure}

	In order to learn the optimal metric space, as in \cite{elad2014_triplet}, we use a CNN, denoted by $\text{CNN}_{\mathcal{D}}$, to learn a mapping function $\mathcal{D}(\cdot)$. The detailed architecture of $\text{CNN}_{\mathcal{D}}$ is shown in Fig.~\ref{fig:cnnD}.
	
	For a random observation triplet $(x_a, x_p, x_n)$, such that $e_{a,p} \in \mathbf{P}$ and $e_{a,n} \in \mathbf{Q}$, we minimize the following loss function to learn the feature map:
	\begin{equation}
	\label{eq:tripletE}
	\begin{medsize}
	Loss_{\mathbf{E}}= \sum_{a,p,n} \big[\alpha_{\mathbf{E}} - \lVert \mathcal{D}(x_a)-\mathcal{D}(x_n)\rVert_2^2 +\lVert \mathcal{D}(x_a)-\mathcal{D}(x_p)\rVert_2^2 \big]_+,
	\end{medsize}
	\end{equation}
	where $\mathcal{D}(\cdot)$ is a CNN-based feature map function, to be learnt, that returns a feature vector corresponding to the input observation, $\alpha_{\mathbf{E}}$ is the minimum margin, and operator $\big[\cdot\big]_+$ is a Rectified Linear Units (ReLU) activation function which is equivalent to $\max(\cdot, 0)$. Let $\mathcal{Z}_{\mathcal{D}}(x)$ be the learnt feature map output at the second to the last layer of $\text{CNN}_{\mathcal{D}}$ (see Fig.~\ref{fig:cnnD}) obtained by minimizing the loss (\ref{eq:tripletE}).
	
	The loss function (\ref{eq:tripletE}) promotes a community structure graph that has relatively small Euclidean distance between the feature maps of vertices connected by the edges in $\mathbf{P}$, and a large distance between the vertices connected by the edges in $\mathbf{Q}$, while keeping a minimum margin $\alpha_{\mathbf{E}}$ between these two distances. 
	
	Since we do not have a priori knowledge of the connectivity of the nodes, we generate the initial graph as a fully connected graph; justification for starting with a fully connected graph is provided in \cite{egilmez2017_graphconstraints}.
	A sparse $\mathbf{E}^0$ minimizes the number of $\mathbf{Q}$ edges by keeping only the connections with $\mathbf{\gamma}^0$ neighbors per individual node. We adopt KNN-graph construction based on (\ref{eq:connectivity}), where optimal maximum number of neighbors $\mathbf{\gamma}^0$ is obtained via grid-search by evaluating classification accuracy of the KNN classifier (denoted by Acc in Fig.~\ref{fig:dynglr})
	using the validation data with the same amount of noisy labels as the training dataset. Note that, as we do not have any prior knowledge of the optimal maximum degree of each individual node, we initially set all $\mathbf{\gamma}^0=\gamma_1=\ldots=\gamma_N$. Once the optimal number of neighbors $\mathbf{\gamma}^0$ is obtained, the resulting graph edges $\mathbf{E}^0$ are used in the following section for pruning edge weights during edge weighting and are updated based on the regularized metric function and the difference between the classifier signal, before and after GLR.
	
	\subsection{W-Net}
	\label{sec:wnet}
	For assigning edge weights $\mathbf{W}^r$ to the graph $\mathbf{G}^r$, we first employ a CNN, denoted by $\text{CNN}_{\mathcal{C}^r}$, to learn a deep metric function. We propose a robust graph-based triplet loss function to better learn feature map $\mathcal{V}^r$, as:
	\begin{equation}
	\label{eq:tripletW}
	\begin{aligned}
	 \medmath{Loss_{\mathbf{W}^r}(\mathcal{V})}=&\medmath{\sum_{\psi_a,\psi_p,\psi_n}\big[\alpha_{\mathbf{W}} - \lVert \mathcal{V}^r(f^r(x_a))-\mathcal{V}^r(f^r(x_n)) \rVert_2^2}\\ &\medmath{\cdot \pi_{(\psi_a,\psi_n|e_{a,n}\in \mathbf{Q})} + \lVert \mathcal{V}^r(f^r(x_a))-\mathcal{V}^r(f^r(x_p)) \rVert_2^2}\\ &\medmath{\cdot \pi_{(\psi_a,\psi_p|e_{a,p} \in \mathbf{P})}
	\big]_+}\\
	{\bf \varPi}^r =&\{\pi_{\psi_i,\psi_j}\}=\{\Theta({\dot{y}_i}^r, {\dot{y}_i}^{r-1}, {\dot{y}_j}^r, {\dot{y}_j}^{r-1})\}.\\
	\end{aligned}
	\end{equation}
	$\Theta$ is an edge attention activation function (see (\ref{eq:edgeatt}) for the particular function we used) that estimates how much attention should be given to each edge and ${\mathbf{Y}^r}=\{\mathbf{\dot{Y}}^r=[-1,1]^M,[-1,1]^{N-M}\}$ is the restored classifier signal obtained via (\ref{eq:glr}) starting from the classifier signal in the previous iteration, $\mathbf{Y}^{r-1}$. $\pi_{\psi_i,\psi_j}$ is the amount of attention, i.e., edge loss weights, assigned to the edge connecting vertices $\psi_i$ and $\psi_j$. 
	Note that $\mathcal{C}^r(\cdot)$ is the feature map learnt by minimizing (\ref{eq:tripletW}). 
	
\begin{figure}[!htbp]
		\centering
		\includegraphics[width=0.85\linewidth]{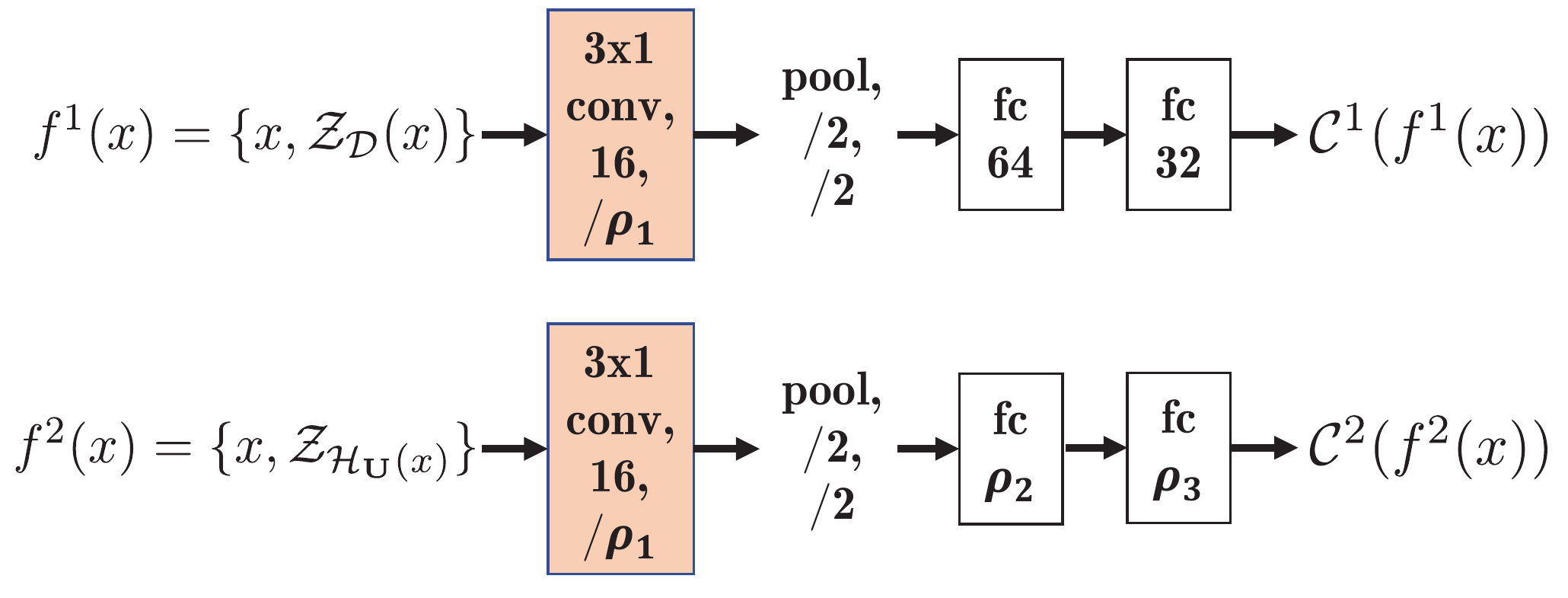}
		\caption{$\text{CNN}_{\mathcal{C}^r}$ neural nets. The stride $\rho_1$ and the number of neurons $\rho_2, \rho_3$ vary depending on the input data (see details in Sec.~\ref{subsec:setup}).}
		\label{fig:cnnC}
	\end{figure}

	The architectures for $r=1$ and $r=2$ are shown in Fig.~\ref{fig:cnnC}.
	Since, at the first iteration $r=1$, we expect many noisy labels, the residual network architecture will be different to the $r>1$ case \cite{he2015_resnet}. 
	
	The architecture presented in Fig.~\ref{fig:cnnC} (top) is used as the feature map $\mathcal{C}^1(\cdot)$, after G-Net, to construct the graph $\mathbf{G^1}=(\mathbf{\Psi},\mathbf{E}^0,\mathbf{W}^1)$ by minimizing $\text{Loss}_{\mathbf{W}^1}(\mathcal{C}^1)$ taking as input undirected graph $\mathbf{G^0}$ learned via G-Net. The input to $\text{CNN}_{\mathcal{C}^1}$ is the concatenated observations $\mathbf{X}$ and ``shallow feature maps" learned via G-Net, i.e., the output of the second to last layer of $\text{CNN}_{\mathcal{D}}$ as presented in Fig.~\ref{fig:cnnD}, denoted by $\mathcal{Z}_{\mathcal{D}}(\mathbf{X})$. 
	
	The $r=2$ architecture is shown in Fig.~\ref{fig:cnnC} (bottom), with observations $\mathbf{X}$ and ``shallow feature maps" learned via U-Net (described in the next subsection) to facilitate the regularization of $\text{CNN}_{\mathcal{C}^2}$ by minimizing $\text{Loss}_{\mathbf{W}^2}(\mathcal{C}^2)$ based on the denoised labels, convolution on both feature maps, denoised classifier signal and their differences across neighbors.
	
	
	Unlike \cite{elad2014_triplet}, we introduce edge attention activation $\Theta$ in (\ref{eq:tripletW}) to dropout some edges with relatively large changes between $\mathbf{\dot{Y}}^r$ and $\mathbf{\dot{Y}}^{r-1}$ via GLR. This helps to focus learning on edges with high confidence given noisy training labels. Therefore, the overall training performance is better than the standard dropout layer approach, which drops out random neuron units in the network. 
	We implement the edge attention activation $\Theta$ and $\Phi$ as:
	\begin{equation}
	 \label{eq:edgeatt}
	\begin{aligned}
	\Phi({ \dot{y}}_i^{r-1}, \dot{y}_i^r)=&\begin{cases}
	1, &\text{if } \lvert \dot{y}_i^{r-1} - \dot{y}_i^r \rvert \leq \varepsilon^r\\
	0, &\text{if } \lvert \dot{y}_i^{r-1} - \dot{y}_i^r \rvert > \varepsilon^r
	\end{cases}\\
	\Theta(\dot{y}_i^{r-1}, \dot{y}_i^r, \dot{y}_j^{r-1},\dot{y}_j^r)=&min(\Phi(\dot{y}_i^{r-1}, \dot{y}_i^r),\Phi(\dot{y}_j^{r-1}, \dot{y}_j^r)),
	\end{aligned}
	\end{equation}
	where threshold $\varepsilon^r$ is used to determine whether a node's label can be trusted and also helps to control the sparsity of edge attention matrix ${\bf \varPi}^r$. 
	That is, if the difference between the signal label in the previous and current iteration is large, this means that the label most likely changed sign (from -1 to +1 or vice versa) and is unreliable in this iteration.
	To reflect the fact that there might be many noisy (unreliable) labels at the start we heuristically set $\varepsilon^1=0.6$ for the first GLR iteration (see the results in Sec.~\ref{sec:simulations}). Since after applying GLR the classification signal is expected to be cleaner, we heuristically set threshold $\varepsilon^2=0.15$ for the second stacked W-Net during training to ensure that we regularize CNNs with less concern about the over-fitting issue introduced by noisy labels.
	
	\subsection{U-Net}
	\label{sec:unet}
	Edge convolution has been proven recently to be a rich feature representation method \cite{zhang2018_edgeconv, wang2018_dyngcnn, zhang2019_edgeconv_action}. We adopt edge convolution in deeper feature map learning, i.e., after GLR $r=1$. That is, given $\mathbf{A}^1$, from the first GLR iteration, each node's feature representation is enhanced by considering observations of both $\mathbf{X}$ and classifier signal $\mathbf{Y}^1$ from its six nearest neighbors (set heuristically), which are most-likely to have the same label.
	\begin{figure}[!htbp]
		\centering
		\includegraphics[width=\linewidth]{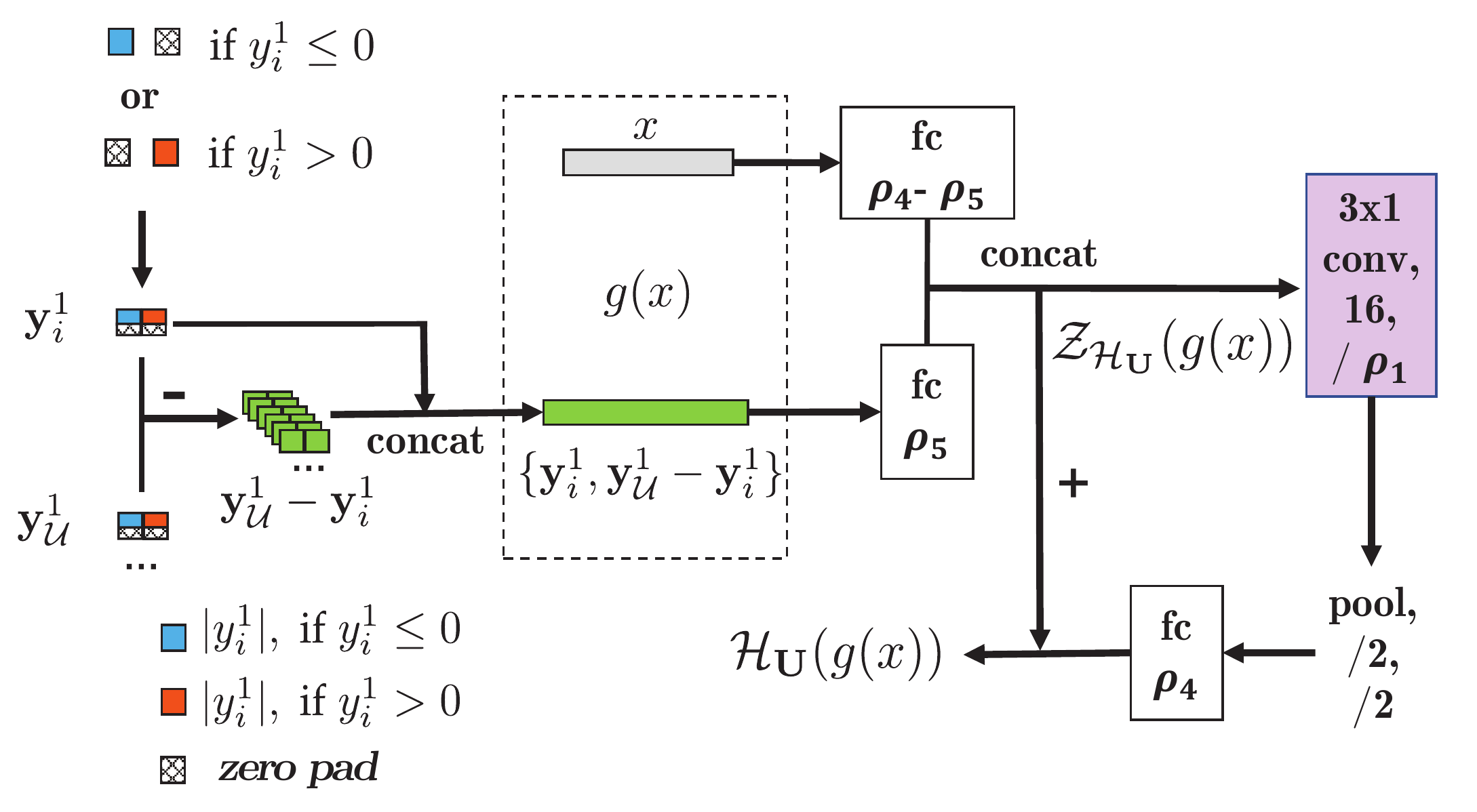}
		\caption{$\text{CNN}_{\mathcal{H}_{\mathbf{U}}}$ neural nets. 
		The stride size $\rho_1$ and the number of neurons $\rho_4, \rho_5$ vary depending on the input data (see details in Sec.~\ref{subsec:setup}).}
		\label{fig:cnnU}
	\end{figure}
	
	Incorporating an additional CNN, denoted by $\text{CNN}_{\mathcal{H}_{\mathbf{U}}}$, shown in Fig.~\ref{fig:cnnU}, we construct a richer feature representation $\mathcal{H}_{\mathbf{U}}(g(x))$ to enhance the graph-based classifier learning with a single input to the network, $g(x)$, comprising $x_i$ and $\{\mathbf{y}^1_i, \mathbf{y}^1_\mathcal{U}-\mathbf{y}^1_i\}$, where $\mathbf{y}^1_i$ denotes a tuple $(y_i, 0)$, if $y_i >0$ or $(0, y_i)$ otherwise. $\mathbf{y}^1_\mathcal{U}$ is a $6 \times 2$ matrix formed by concatenating $\mathbf{y}^1_i$ with the nearest six neighboring nodes. Finally,  
	$\mathbf{y}^1_\mathcal{U}-\mathbf{y}^1_i$ is obtained by subtracting each row of $\mathbf{y}^1_\mathcal{U}$ by $\mathbf{y}^1_i$.
	
	Graph edge $\mathbf{E}^{r-1}$ is updated by (\ref{eq:graphupdate}) based on the learnt regularized feature map $\mathcal{H}_{\mathbf{U}}(\cdot)$ in order to better reflect the node-to-node correlation. The new edge matrix $\mathbf{E}^1$ and the denoised classifier signal $\mathbf{Y}^1$ are then used in the second graph-based classifier iteration.

	Though we can continue iterating between W-Net and U-Net, in our practical implementation, only two iterations are performed, to reduce computation complexity, since heuristically we did not observe improvement after $r=2$ iterations.

	\section{Simulations}
	\label{sec:simulations}
	In this section, we present our simulation results, including the ablation study, visualization results, and comparison of the performance against different, classic and state-of-the-art classifiers, under different label noise levels. 
	
	\subsection{Simulation setup: Datasets, Benchmarks, Parameters and Performance Measure}
	\label{subsec:setup}
	
	\subsubsection{Datasets}
	We select three binary-class datasets from Knowledge Extraction based on Evolutionary Learning dataset (KEEL) \cite{keel2010} that vary in the number and type of features; these sets are, from low dimensional feature sets to higher ones: (1) \textbf{Phoneme}: contains nasal (class 0) and oral sounds (class 1), with 5404 instances (frames) described by 5 phonemes of digitized speech. (2) \textbf{Magic}: contains images generated by primary gammas (class 0) from the images of hadronic showers initiated by cosmic rays in the upper atmosphere (class 1), where 19020 instances are generated for simulation using the imaging technique, with each instance containing 10 attributes to characterize simulated images. (3) \textbf{Spambase}: to determine whether an email is spam (class 0) or not (class 1), with 4597 email messages summarized by 57 particular words or characters.
	
	\subsubsection{Benchmark classifiers}
	We compare the proposed network against 10 different classification methods: (1) SVM with radial basis function kernel (SVM-RBF) (2) a classical CNN, consisting of two CNN blocks and two fully connected layers afterwards, where each CNN block has a convolution layer, a max pooling layer and one dropout layer (3) a graph CNN with multiple graph construction blocks, where each block constructs a KNN graph based on multiple graph structures learnt via edge convolution; batch normalization with decay is used (called DynGraph-CNN \cite{wang2018_dyngcnn}) (4) a KNN classifier using CNN-based deep metric learning (used CNN is the same as $\text{CNN}_{\mathcal{D}}$) \cite{elad2014_triplet} (called DML-KNN) (5) a rank-sampling \cite{wang2019_rankingclean} based KNN classifier using CNN-based deep metric learning (CNN used is same as in DynGraph-CNN), where sampling is performed on the training set by calculating the resulting classification accuracy using randomly sampled samples. We use the top 480 training samples with relatively high classification accuracy in the validation set. During inference, 480 selected training samples are divided into 6 equal-size batches by stratified random sampling and the predictions using each batch are averaged to obtain the final decision (6) label noise robust SVM with RBF kernel \cite{biggio2011_lnrsvm} (LN-Robust-SVM-RBF) (7) a graph-based classifier with negative edge weights assigned between the centroid sample pairs and between the boundary sample pairs (named Graph-Hybrid \cite{gene2018_negglrgu}) (8) a CNN network (same as in DynGraph-CNN) trained by savage loss (called CNN-Savage \cite{masnadi2009_savage}) (9) a CNN network (same as in DynGraph-CNN) trained by bootstrap-hard loss (called CNN-BootStrapHard \cite{reed2014_bootstrap}) (10) a CNN network (same as in DynGraph-CNN) trained via dimensionality-driven learning strategy (called CNN-D2L \cite{ma2018_d2l}). 
	Note that Classifiers (1)-(4) are classical methods for normal, `noise-free', conditions, and methods (5)-(9) are proposed to avoid overfitting under noisy training labels. All CNN-based methods adopt $l_2$ regularization for each layer. Similar to Benchmark (5), we also adopt rank-sampling technique on the training set to select trusted samples that will further facilitate the predictive performance and consistency of our model, denoted by `s' appended to the model name.
	
	\subsubsection{Ablation study}
	\label{sec:ablationstudy}
	To understand how different components of our proposed architecture affect the results, we perform an ablation study by removing some components. The resulting architectures are denoted by DynGLR-G-number, where `G' refers to Graph generation and  `1’ refers to edge weighting, `2’ to GLR, and `3’ graph update. That is, the following variants of the proposed scheme are compared: (1) DynGLR-G-2: we import the unweighted graph $\mathbf{G}^0$ generated by G-Net (see Fig.~\ref{fig:generator}) into GLR for classification. (2) DynGLR-G-12: we assign weights to the unweighted graph $\mathbf{G}^0$ via an adaptive Gaussian kernel function (see (Eq.~\ref{eq:edgeweight})); the resulting undirected and weighted graph is then used to perform node classification via GLR. (3) DynGLR-G-1232: we update the graph edge sets by considering the neighbors of each node with denoised classifier signal and observed feature maps (see Fig.~\ref{fig:graphupdate}); the resulting unweighted graph is then used for classification. (4) DynGLR-G-12312: we reassign weights to the updated unweighted graph to effectively perform classification; we perform rank-sampling for all architectures to evaluate the benefits, denoted by `s' appended to the name of each proposed architecture.
	
		\subsubsection{Simulation setup, performance measure and parameters}
	We design our experiments by splitting each dataset into training, validation and testing sets with 40\%, 20\%, 40\% of instances, respectively, with balanced class distribution. To evaluate the robustness of different classification methods against label noise, we randomly sample subsets of instances from both training and validation sets and reverse their labels. Classification error rates are measured by running 20 experiments per label noise level (0\% to 25\%). We use the same random seed setting across all classification methods and remove all duplicated instances to ensure a fair comparison. 
	
	Hyper-parameters used for each experiment are obtained from the validation sets by grid search. All used parameters are listed in Table~\ref{tab:params}.
	\begin{table}[!htbp]
		\centering
		\caption{Parameters for the proposed architectures. $x\Rightarrow y$ means that the learning rate decreases linearly from $x$ to $y$ with the epoch number.}
		\label{tab:params}
		\begin{tabu} to \linewidth{|c|X[c]|X[c]|X[c]|}
			\hline
			Hyper-parameters & \textit{\textbf{Phoneme}} & \textit{\textbf{Magic}} & \textit{\textbf{Spambase}} \\ \hline
			$\rho_1,\rho_2,\rho_3,\rho_4,\rho_5$ & 1,256,64,256,6 & 1,128,32,128,4 & 2,32,32,64,6 \\ \hline
			G-Net learning rates & 0.02$\Rightarrow$0.01 & 0.02$\Rightarrow$0.01 & 0.02$\Rightarrow$0.01 \\ \hline
			G-Net epochs & 160 & 160 & 60 \\ \hline
			W-Net($r$=1) learning rates & 0.02$\Rightarrow$0.01 & 0.02$\Rightarrow$0.01 & 0.02$\Rightarrow$0.012 \\ \hline
			W-Net($r$=1) epochs & 320 & 320 & 80 \\ \hline
			U-Net learning rates & 0.002$\Rightarrow$0.001 & 0.002$\Rightarrow$0.001 & 0.002$\Rightarrow$0.001 \\ \hline
			U-Net epochs & 120 & 180 & 100 \\ \hline
			W-Net($r$=2) learning rates & 0.01$\Rightarrow$0.002 & 0.01$\Rightarrow$0.002 & 0.02$\Rightarrow$0.01 \\ \hline
			W-Net($r$=2) epochs & 60 & 40 & 40 \\ \hline
		\end{tabu}
	\end{table}
	
	To guarantee the solution $\mathbf{Y}^r$ to (\ref{eq:glr}) is numerically stable, we heuristically set conditional number $\kappa=60$ and $\mu^r=0.67\mu^r_{max}$ in all our experiments. We use the distance margin $\alpha_{\mathbf{E}}=\alpha_{\mathbf{W}}=10$ in both (\ref{eq:tripletE}) and (\ref{eq:tripletW}). For each epoch, we use batch size of 16, each batch comprising 80 labeled instances from training set and 20 unlabeled instances from validation set; thus we randomly select $16 \cdot 100$ instances. This results in 16 graphs to regularize training per epoch.
	
	All CNNs are learnt by ADAM optimizer, classification accuracy and classifier signal changes are used as metric for rank-sampling for further improving the predictive performance.
	
	\subsection{Results and Discussion}
	\label{subsec:results}
	This section is organized as follows.
	As part of the ablation study, Subsecs.~\ref{ss:1Eval} and \ref{ss:3Eval} evaluate the ability of the proposed schemes to clear the noisy labels and observation samples during the training phase, respectively. We analyze the sensitivity of hyper-parameters that affect the regularization performance in Subsec.~\ref{ss:2Hyper}. We show the effectiveness of our iterative graph update scheme in Subsec.~\ref{ss:4Graph} by visualizing the learnt underlying graph in spectral domain. To analyze the impact of different components on the classification accuracy, we show the classification error rates in Subsec.~\ref{ss:5Class} and discuss the findings of our ablation study during the testing phase and show comparison with state-of-the-art schemes.
	
	\subsubsection{Evaluating graph update block}
	\label{ss:1Eval}
	\begin{figure}[!htbp]
		\centering
		\includegraphics[width=0.95\linewidth]{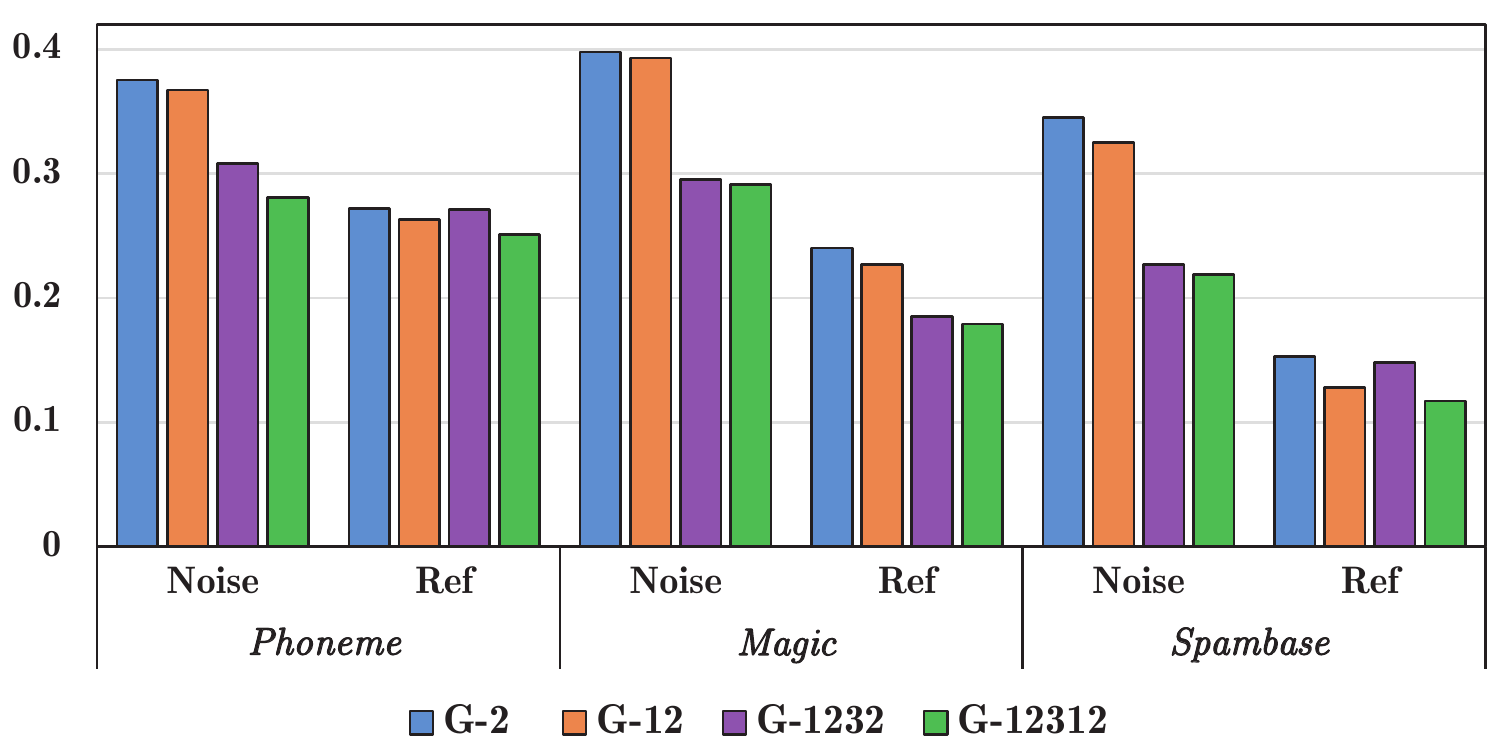}
		\caption{Mean Edge Weight Proportion $\varrho$ for the proposed DynGLR-Nets for all three datasets when 25\% labels used for training are wrong (denoted by `Noise') or when all training labels are correct (denoted by `Ref' without use of GLR). G-2, G-12, G-1232, and G-12312 schemes are described in the previous subsection.}
		\label{fig:dglc_ew_visual}
	\end{figure}
	
	First, we evaluate ability of the graph update block to clear noisy labels. We use the mean edge weight proportion measure defined as:
	\begin{equation}
	\varrho= \frac{\sum_{p,n}^{N} w_{p,n}}{\sum_{i,j}^{N} \mathbbm{1} (w_{i,j}>0)},
	\label{equ:mewp}
	\end{equation}	
	where $\psi_p$ and $\psi_n$ are two nodes with the opposite labels, $w_{p,n}$ is the weight of the edge $e_{p,n}$ and $\mathbbm{1}(c)$ is an operator that returns 1 if condition $c$ is fulfilled, and 0 otherwise. The results are shown in Fig.~\ref{fig:dglc_ew_visual}, which shows that, for all datasets, the number of connections between nodes with opposite labels decreases with iterations and becomes similar to that without any noise, indicating that the graph update manages to restore the noisy labels.
	
	\begin{figure}[!htbp]
		\centering
		\captionsetup[subfigure]{labelformat=empty, justification=centering}
		\begin{tabu} to \linewidth{@{\hskip 2pt}c@{\hskip 4pt}c@{\hskip 2pt}}
			~~~~Clean Vertices &~~~~Noisy Vertices\\
			\includegraphics[align=c,width=0.48\linewidth]{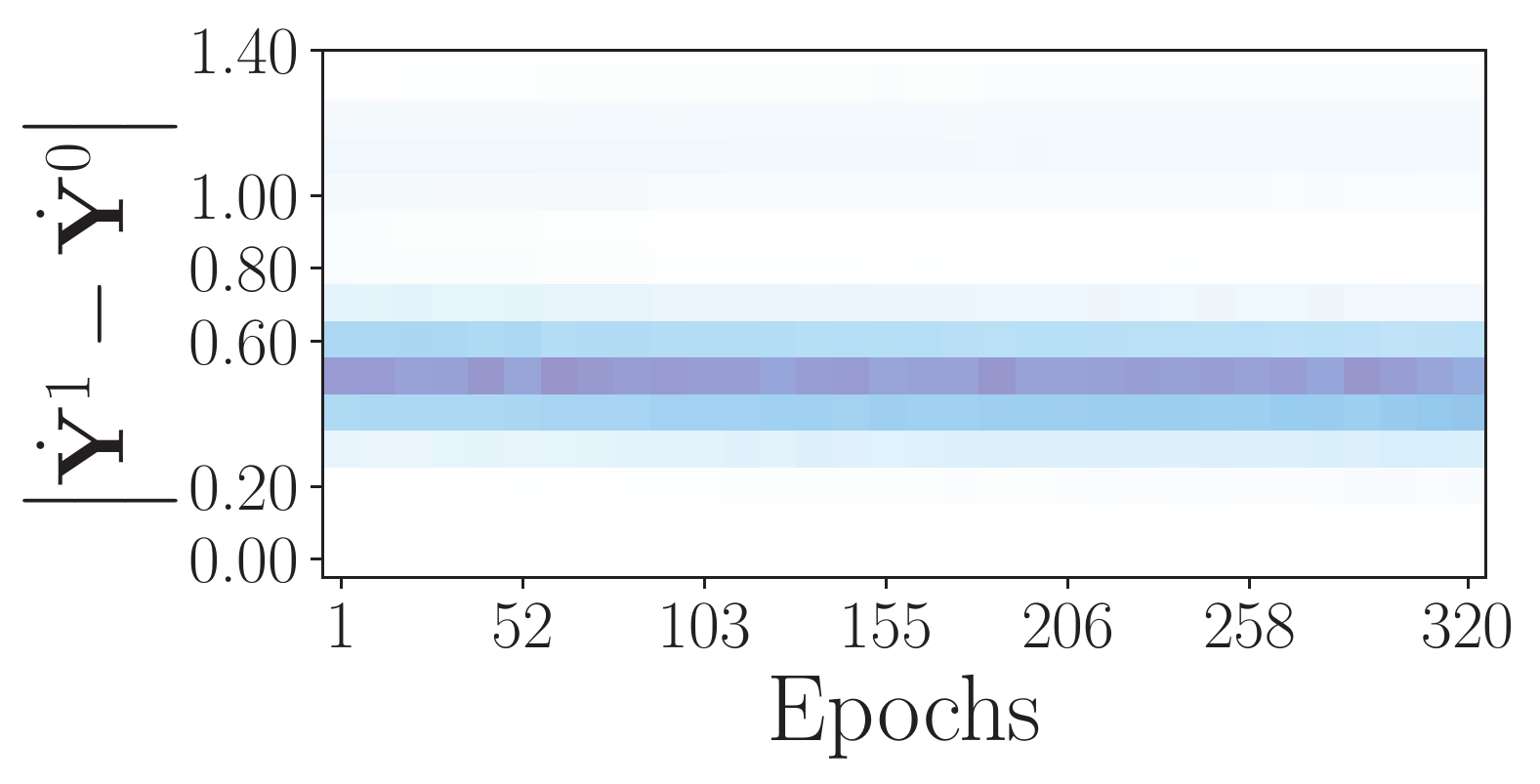}&
            \includegraphics[align=c,width=0.48\linewidth]{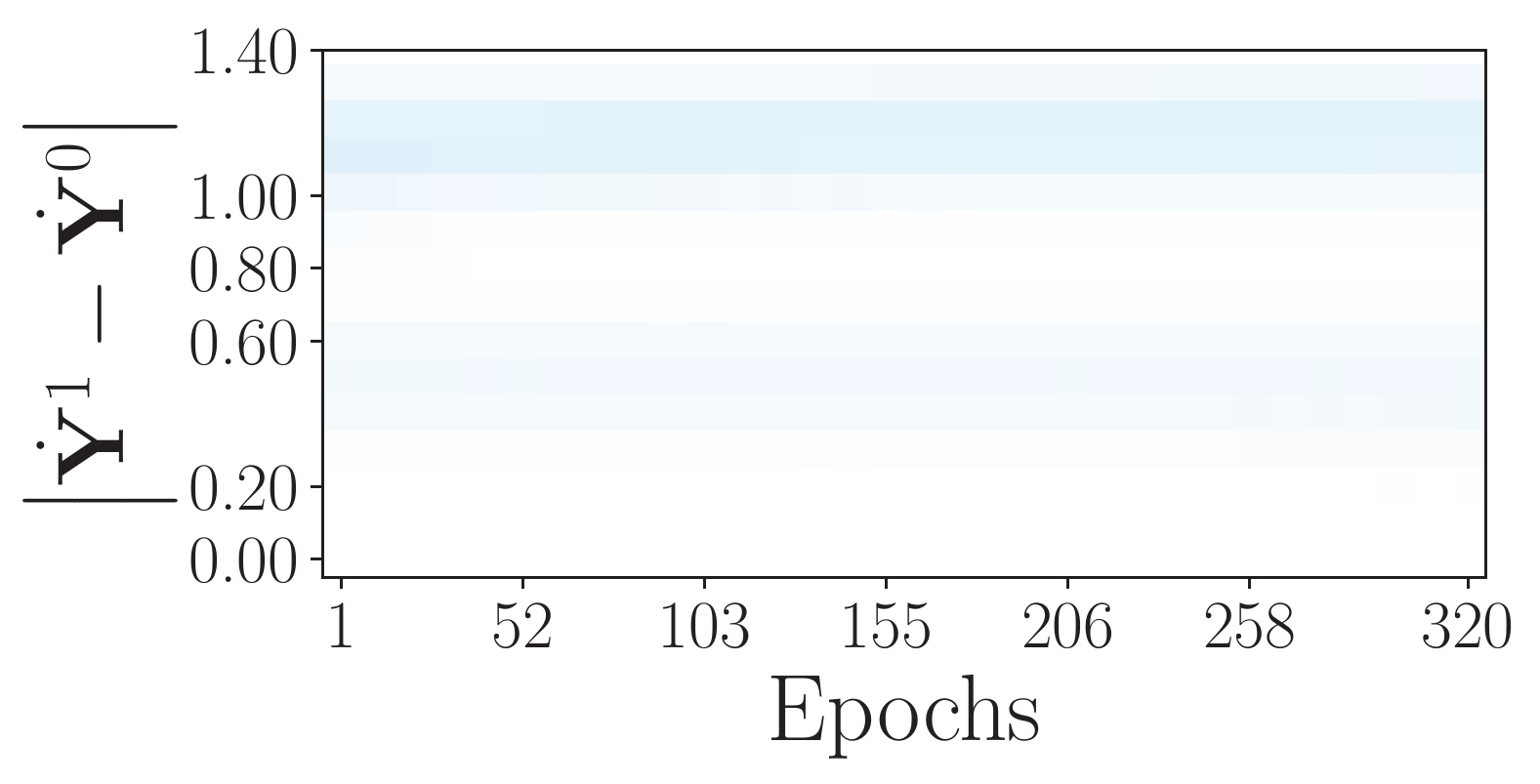}\\
			\includegraphics[align=c,width=0.48\linewidth]{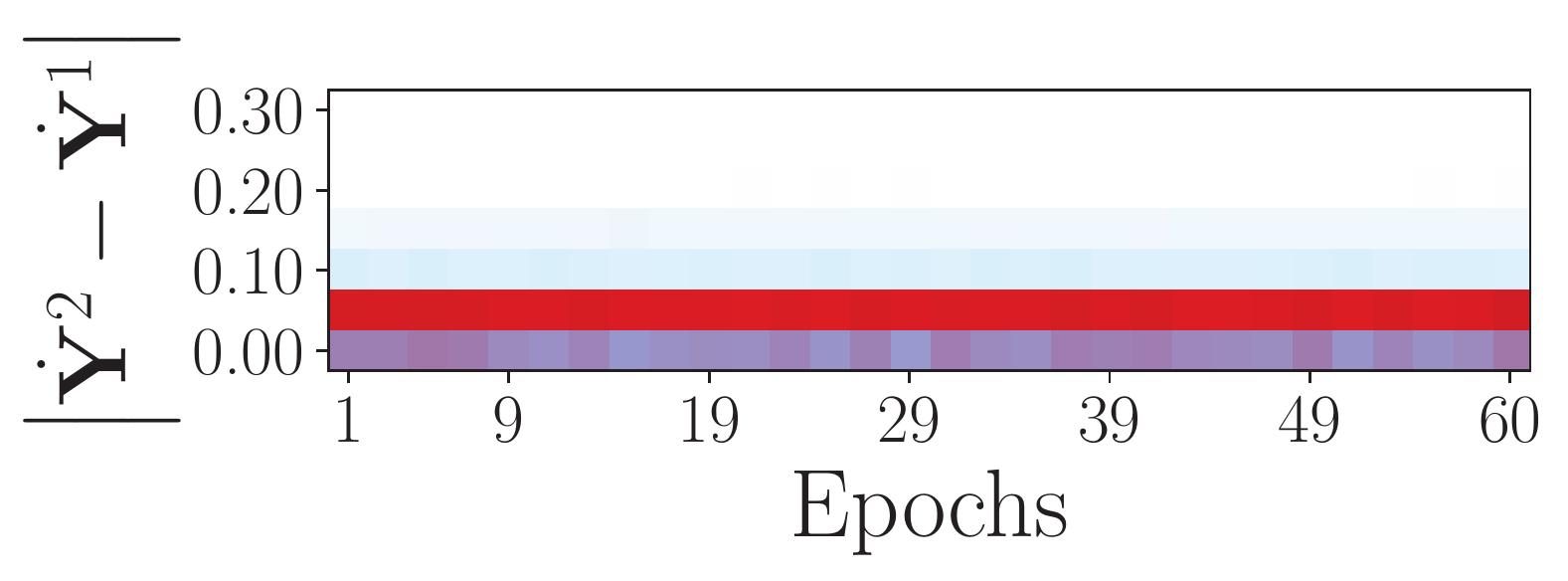}&
			\includegraphics[align=c,width=0.48\linewidth]{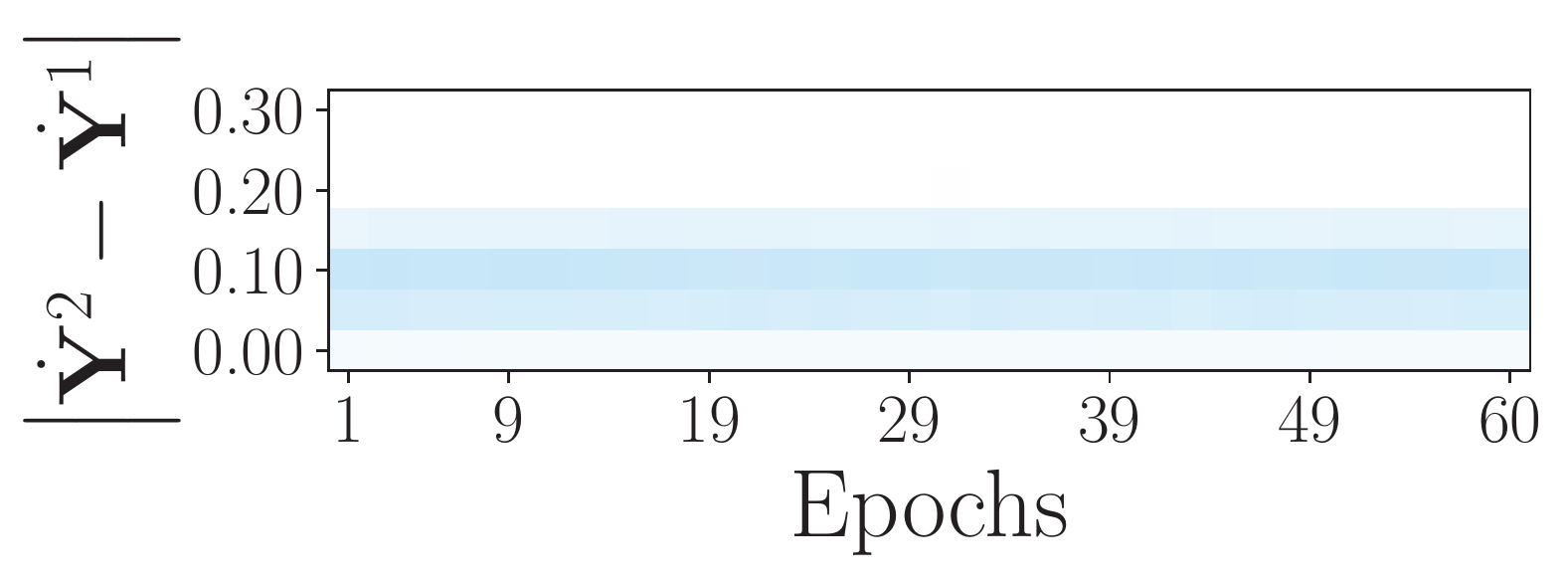}\\
			\multicolumn{2}{c}{\includegraphics[align=c,width=0.6\linewidth]{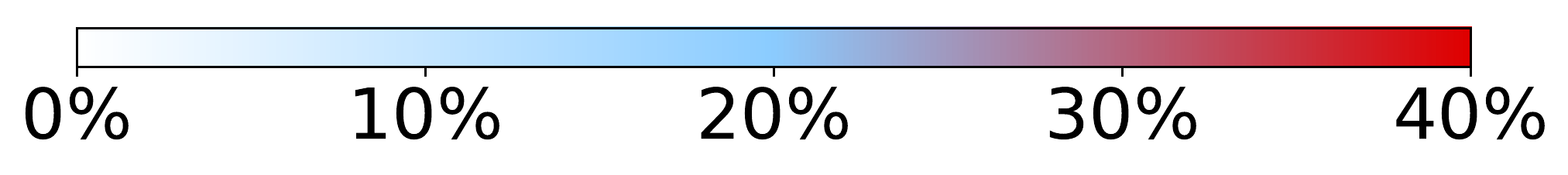}}
		\end{tabu}
		\caption{Classifier signal changes $\lvert \dot{\mathbf{Y}}^{r-1}-\dot{\mathbf{Y}}^r \rvert$ density visualization for \textit{\textbf{Phoneme}} dataset after the first (top row) and second (bottom row) GLR iteration during the training period. Vertices with clean/incorrect labels are shown in the first/second column. 
		The intensity of the classifier signal changes across all experiments are represented through colormaps.}
		\label{fig:phoneme_epsilon}
	\end{figure}
	
	In Eq.~(\ref{eq:edgeatt}) we use a threshold to distinguish reliable nodes from unreliable nodes. In order to evaluate the used approach and to set thresholds, we show the change of the classifier signal $\lvert \dot{\mathbf{Y}}^{r-1}-\dot{\mathbf{Y}}^r \rvert$ during the first two iterations in Fig.~\ref{fig:phoneme_epsilon}. 
	We can see that when all the labels are clean (left column) the difference between the signals before and after GLR is mainly below 0.6 and 0.15, in the first and the second iteration, respectively. Thus, by setting the thresholds at $\varepsilon^{1} \approx 0.6$ and $\varepsilon^{2} \approx 0.15$ for the first and the second GLR iteration, respectively, we can distinguish the vertices with potentially noisy labels. Similar observations are made for the \textit{\textbf{Spambase}} and \textit{\textbf{Magic}} datasets.
	
	
	Heuristically, we observed that as more GLR iterations are performed, the overlap between the clean and noisy vertices distribution of classifier signal changes is high, resulting in reduced ability to use thresholding for distinguishing if a node is sufficiently cleaned. That is why in all our experiments, to reduce complexity, we perform the graph update only after the first iteration, i.e., for $r$=1. We next assess sensitivity of the network to threshold values.
	
	\begin{figure}[!htbp]
		\centering
		\subfloat[GLR ($r$=1)]{\includegraphics[width=0.48\linewidth]{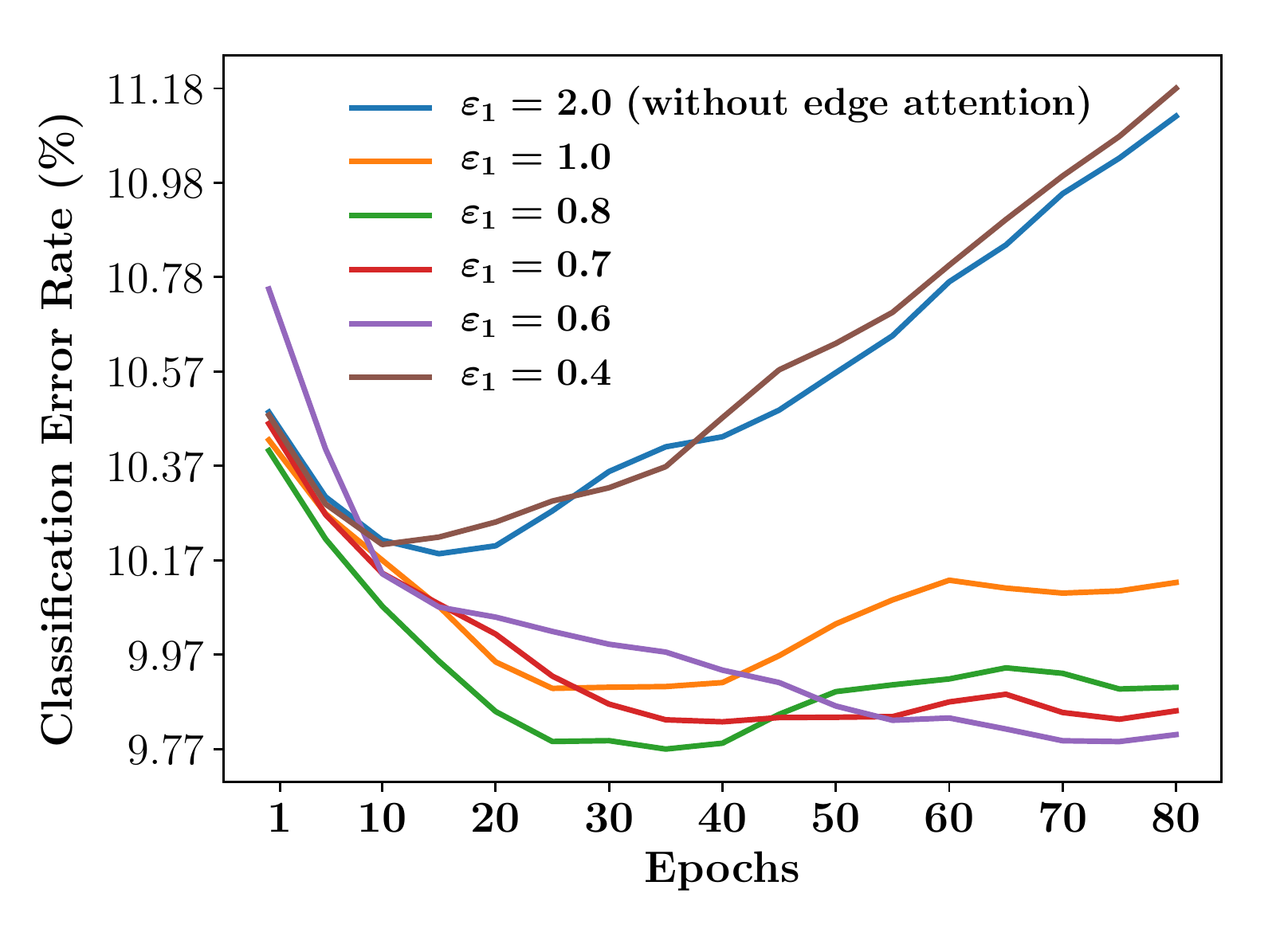}%
        \label{fig:spambase_epsilon1}%
        }
        \subfloat[GLR ($r$=2)]{\includegraphics[width=0.48\linewidth]{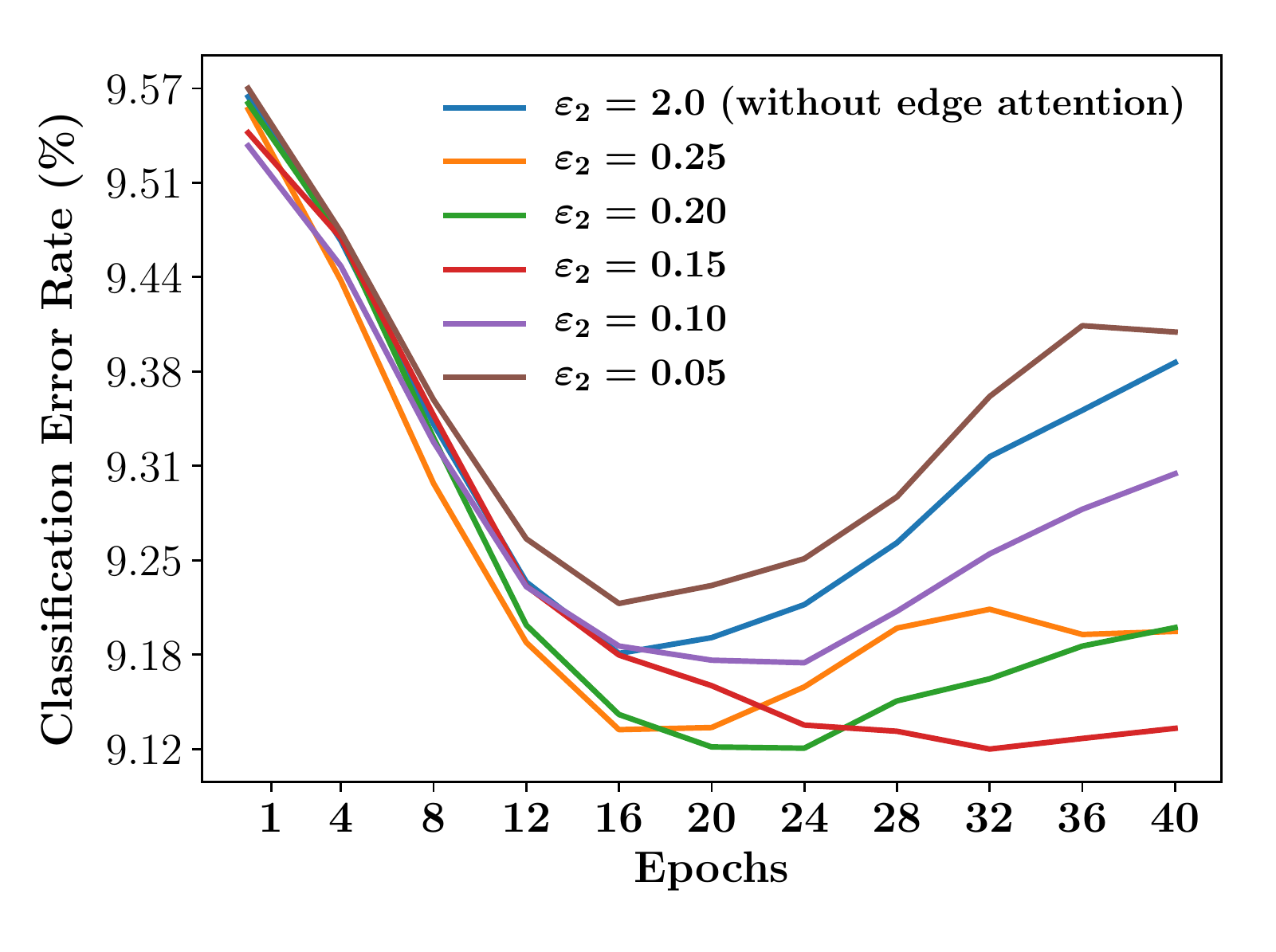}%
        \label{fig:spambase_epsilon2}%
        }
		\caption{Classification Error Rate (\%) for \textit{\textbf{Spambase}} dataset using different $\varepsilon^{1,2}$ in DynGLR-G-12312.}
		\label{fig:spambase_epsilon}
	\end{figure}
	
	\subsubsection{Hyper-parameter sensitivity}
	\label{ss:2Hyper}
	We use attention activation (Eq.~\ref{eq:edgeatt}) to detect the label of vertex $\psi_i$ as possibly noisy if $\Phi(\dot{y_i}^{r-1}, \dot{y_i}^r)=1$. To analyze the sensitivity of hyper-parameters $\varepsilon^1$ and $\varepsilon^2$ (thresholds) in (Eq.~\ref{eq:edgeatt}), we show the classification error rate for the $\textbf{\textit{Spambase}}$ dataset during training using different values of $\varepsilon^{1}$ and $\varepsilon^{2}$ in Fig.~\ref{fig:spambase_epsilon}. We observe that the  thresholds $\varepsilon^1=0.6,\varepsilon^2=0.15$, from the density visualization of classifier signal changes $\lvert \dot{\mathbf{Y}}^{r-1}-\dot{\mathbf{Y}}^r \rvert$ of Fig.~\ref{fig:phoneme_epsilon}, are appropriate to improve the regularization of CNNs. We also show that the classification error rate reduces from first GLR iteration to second GLR iteration.
	
	\subsubsection{Evaluating Rank-sampling}
	\label{ss:3Eval}
	
		\begin{figure}[!htbp]
		\centering
		\includegraphics[width=0.95\linewidth]{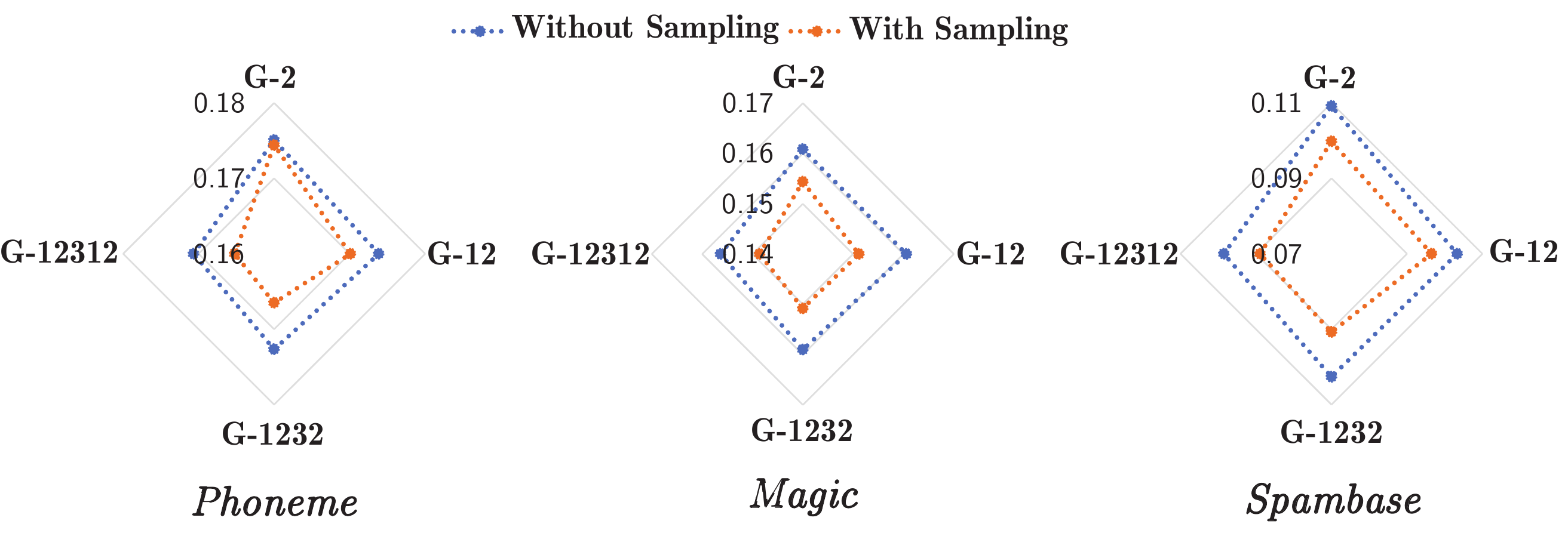}
		\caption{The mean noise level of the training labels after GLR is performed in all proposed DynGLR-Nets when 25\% labels used for training are incorrect.}
		\label{fig:noiselevel_radarchart}
	\end{figure}
	
	To evaluate the denoising effects of GLR, in Fig.~\ref{fig:noiselevel_radarchart}, we show the mean noise level of the training labels after GLR is performed across all our proposed architectures with and without sampling. It is clear from Fig.~\ref{fig:noiselevel_radarchart}, for all three datasets, that the mean noise level is lower with rank-sampling than without. This confirms that with rank-sampling, one can further reduce the effects of noisy training labels by dropping out the less reliable training samples.

\subsubsection{Graph spectrum visualization}
	\label{ss:4Graph}
	
\begin{figure}[!htbp]
		\centering
		\captionsetup[subfigure]{labelformat=empty, justification=centering}
		\begin{tabu} to \linewidth{@{\hskip 2pt}c@{\hskip 2pt}c@{\hskip 2pt}}
			~~~~Unweighted &~~~~Weighted\\
			\includegraphics[align=c,width=0.48\linewidth]{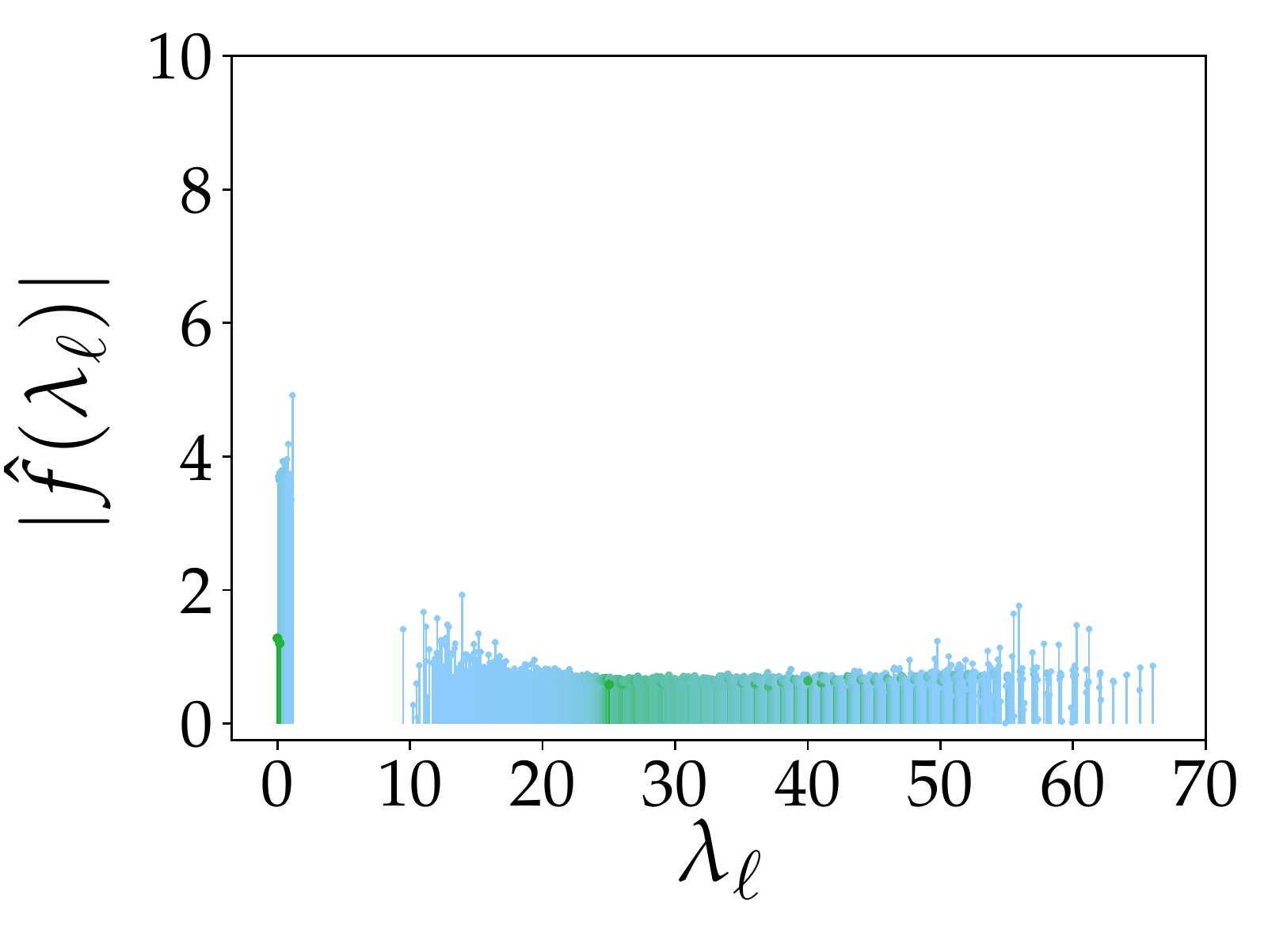}&
            \includegraphics[align=c,width=0.48\linewidth]{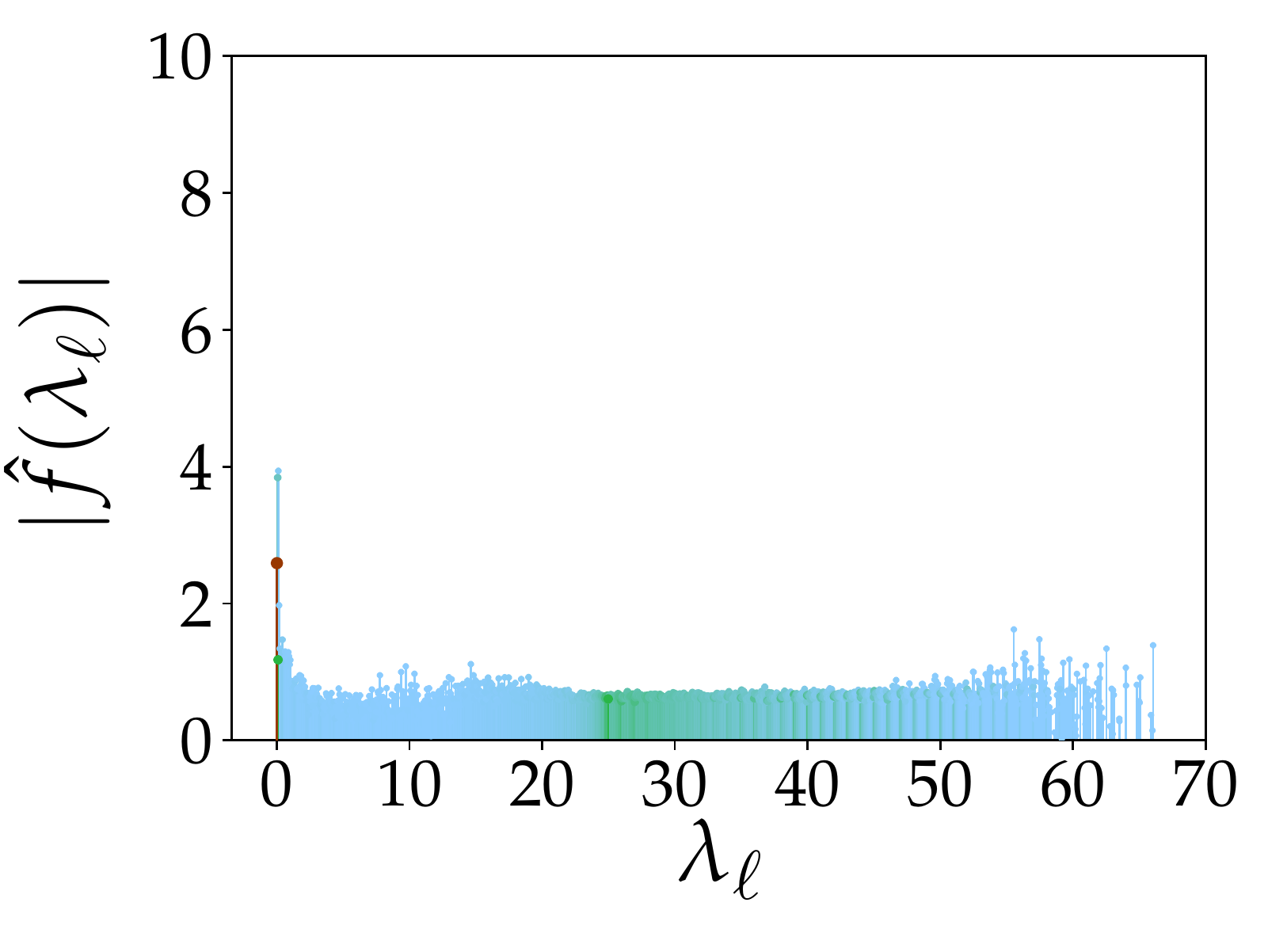}\\
			\includegraphics[align=c,width=0.48\linewidth]{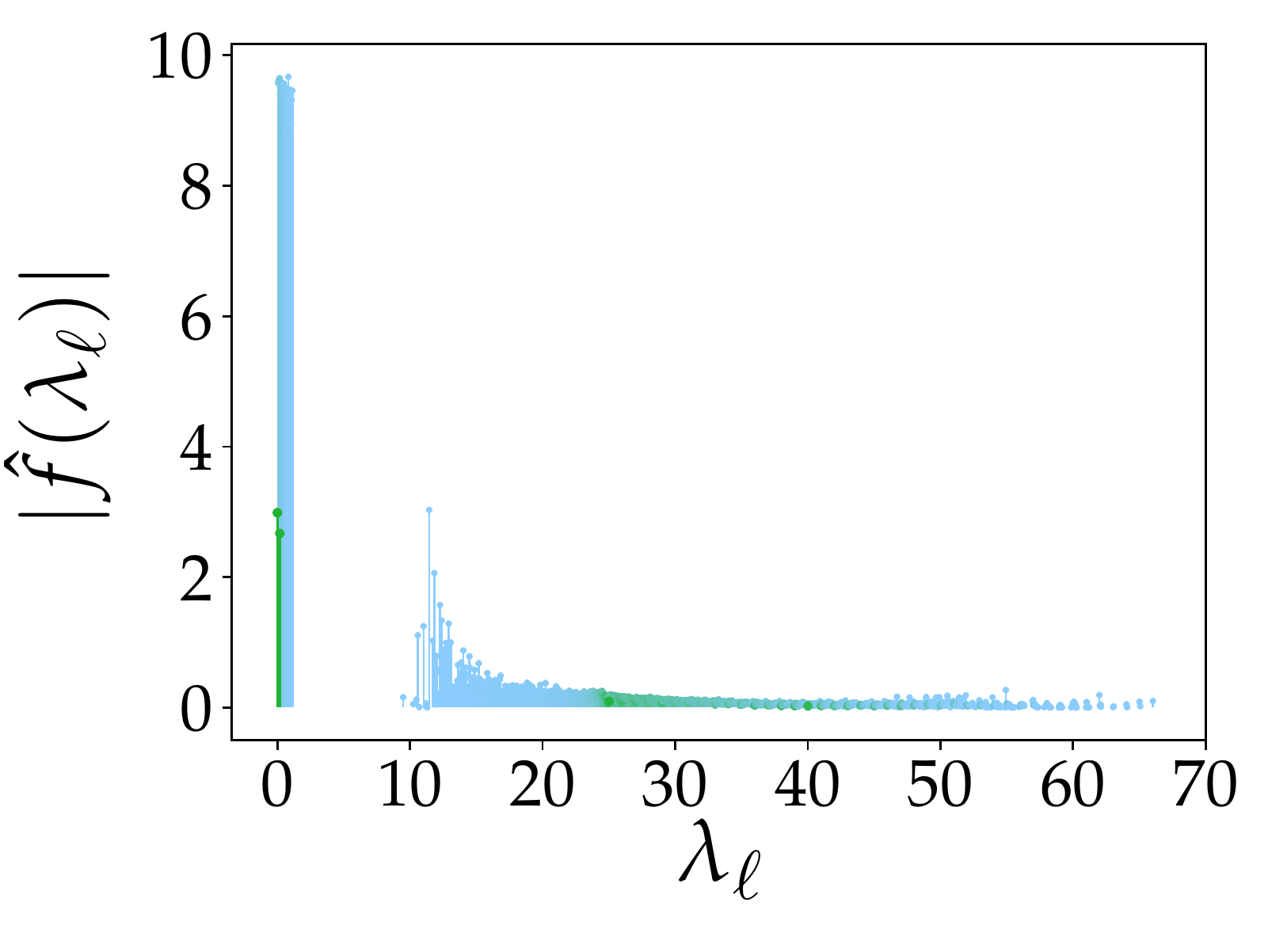}&
			\includegraphics[align=c,width=0.48\linewidth]{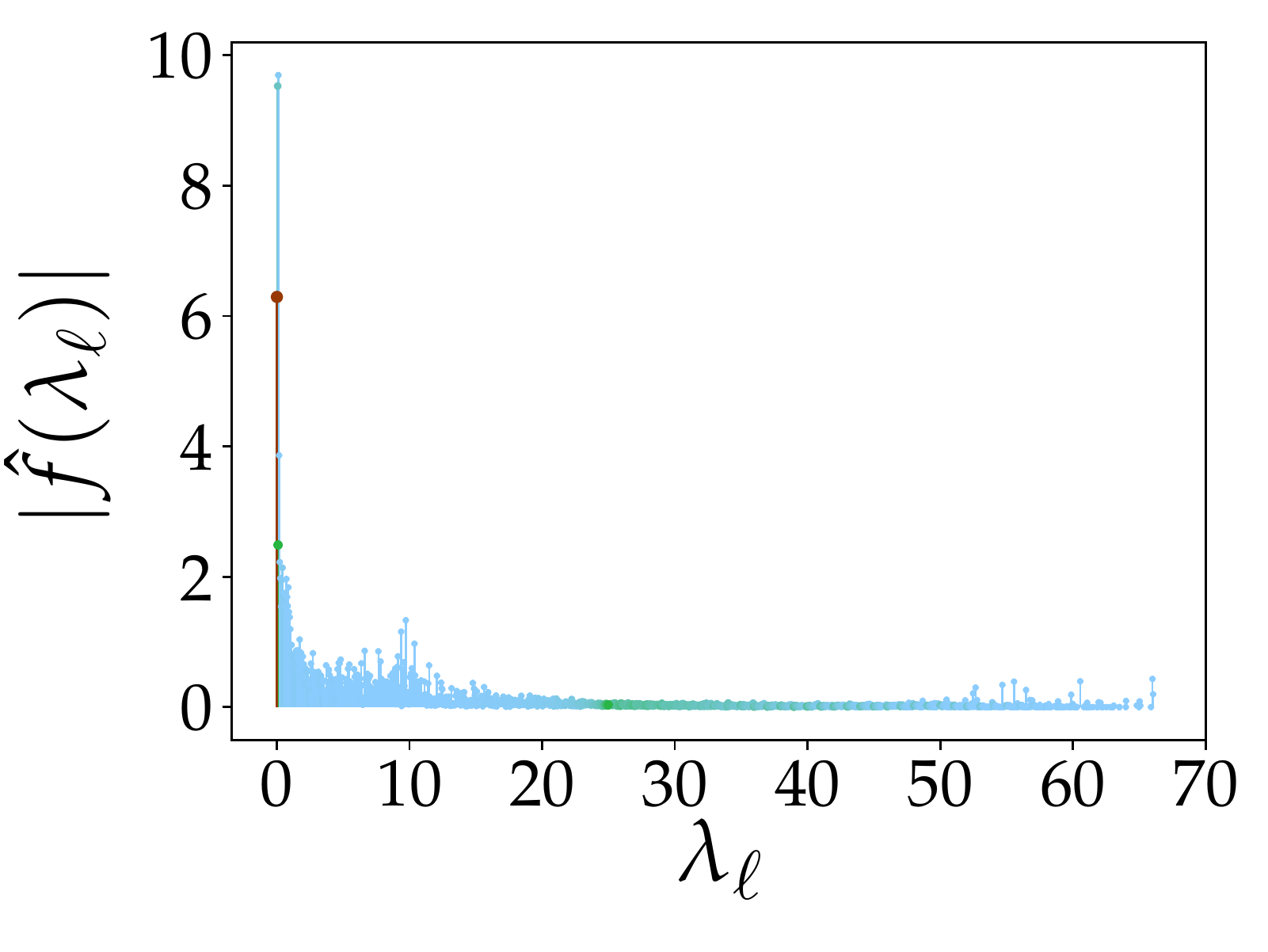}\\
			\includegraphics[align=c,width=0.48\linewidth]{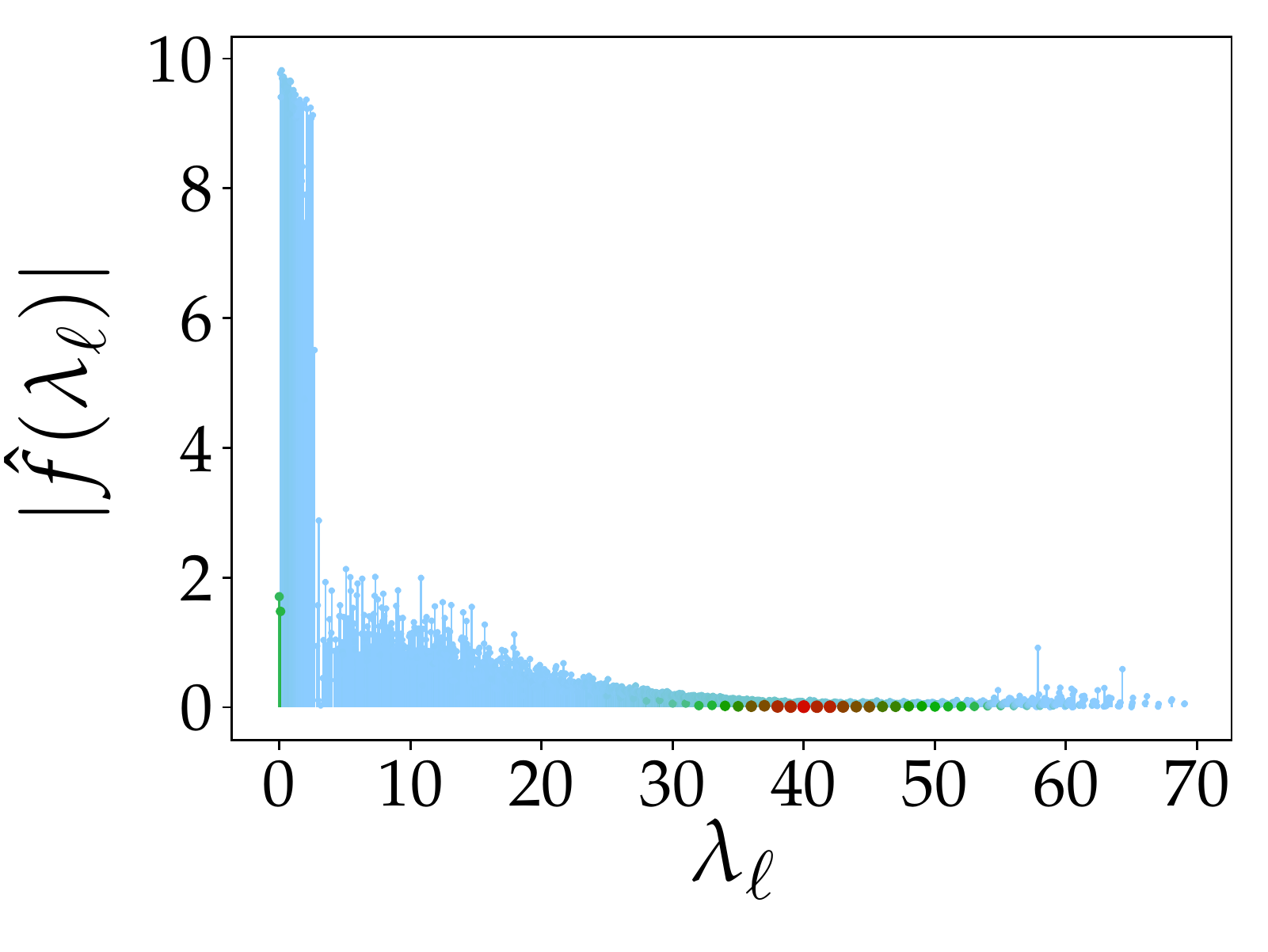}&
			\includegraphics[align=c,width=0.48\linewidth]{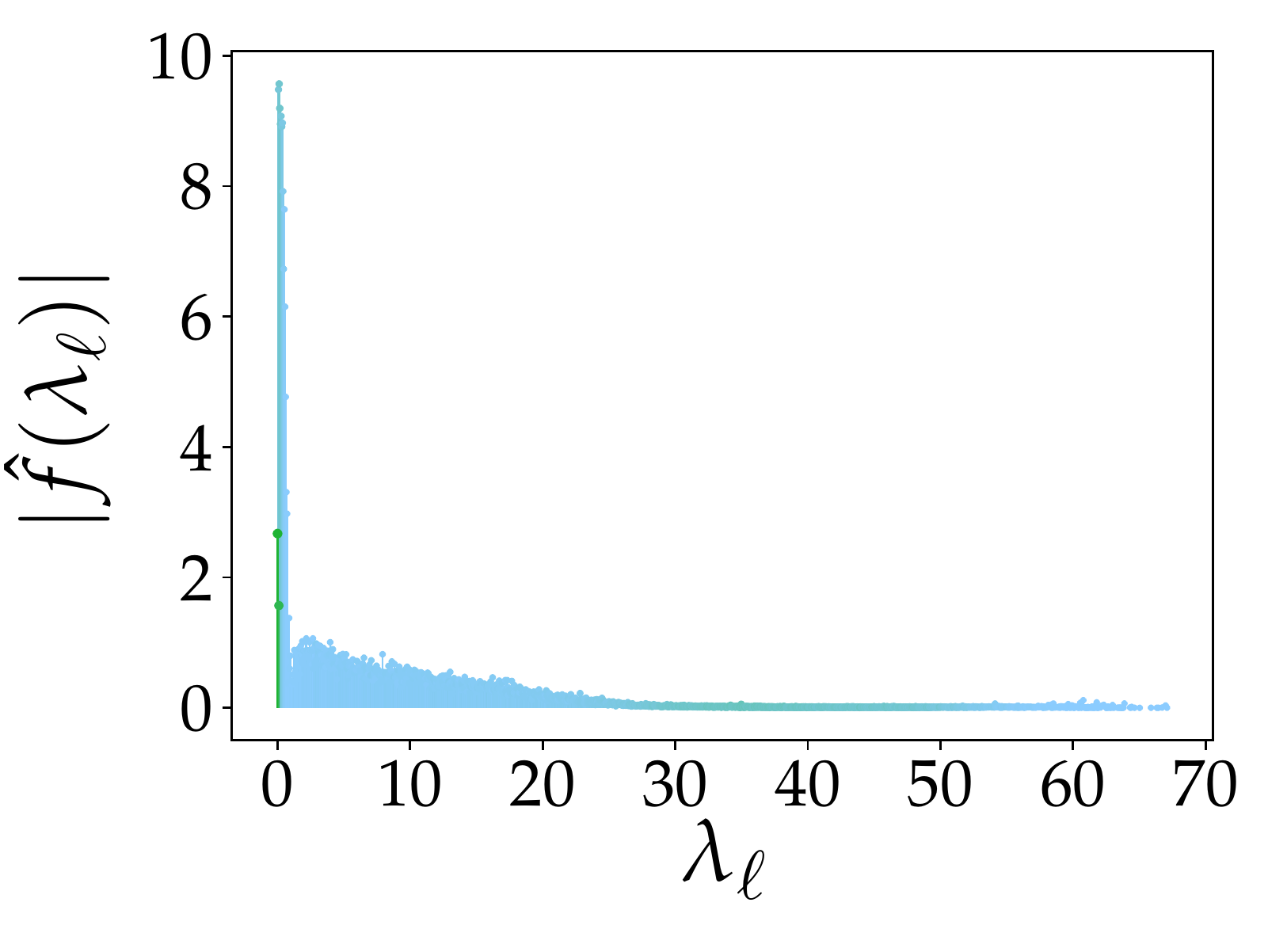}\\
			\multicolumn{2}{c}{\includegraphics[align=c,width=0.7\linewidth]{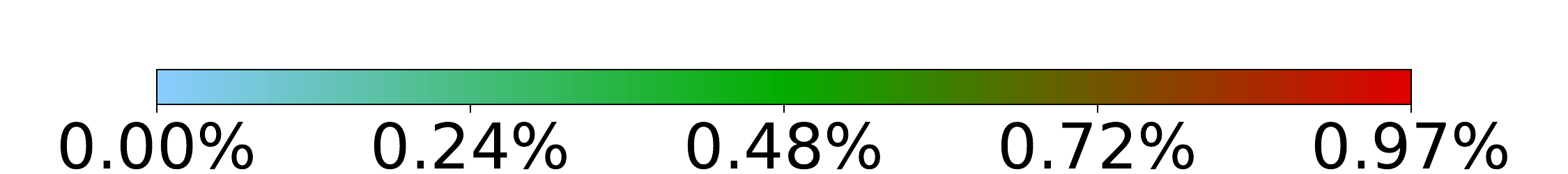}}
		\end{tabu}
		\caption{The magnitude of the Graph Fourier Transform coefficients for \textit{\textbf{Phoneme}} Dataset. The density of each Eigenvalue $\lambda_\ell$ across all experiments on the testing sets is represented through colormaps. Top row shows the result after initialization and before GLR (G-Net output) and the second and third row show the result after the first ($r=1$) and the second iteration ($r=2$), respectively. 
		}
		\label{fig:phoneme_gft}
	\end{figure}

\begin{figure}[!htbp]
		\centering
		\captionsetup[subfigure]{labelformat=empty, justification=centering}
		\begin{tabu} to \linewidth{@{\hskip 2pt}c@{\hskip 2pt}c@{\hskip 2pt}}
			~~~~Unweighted &~~~~Weighted\\
			\includegraphics[align=c,width=0.48\linewidth]{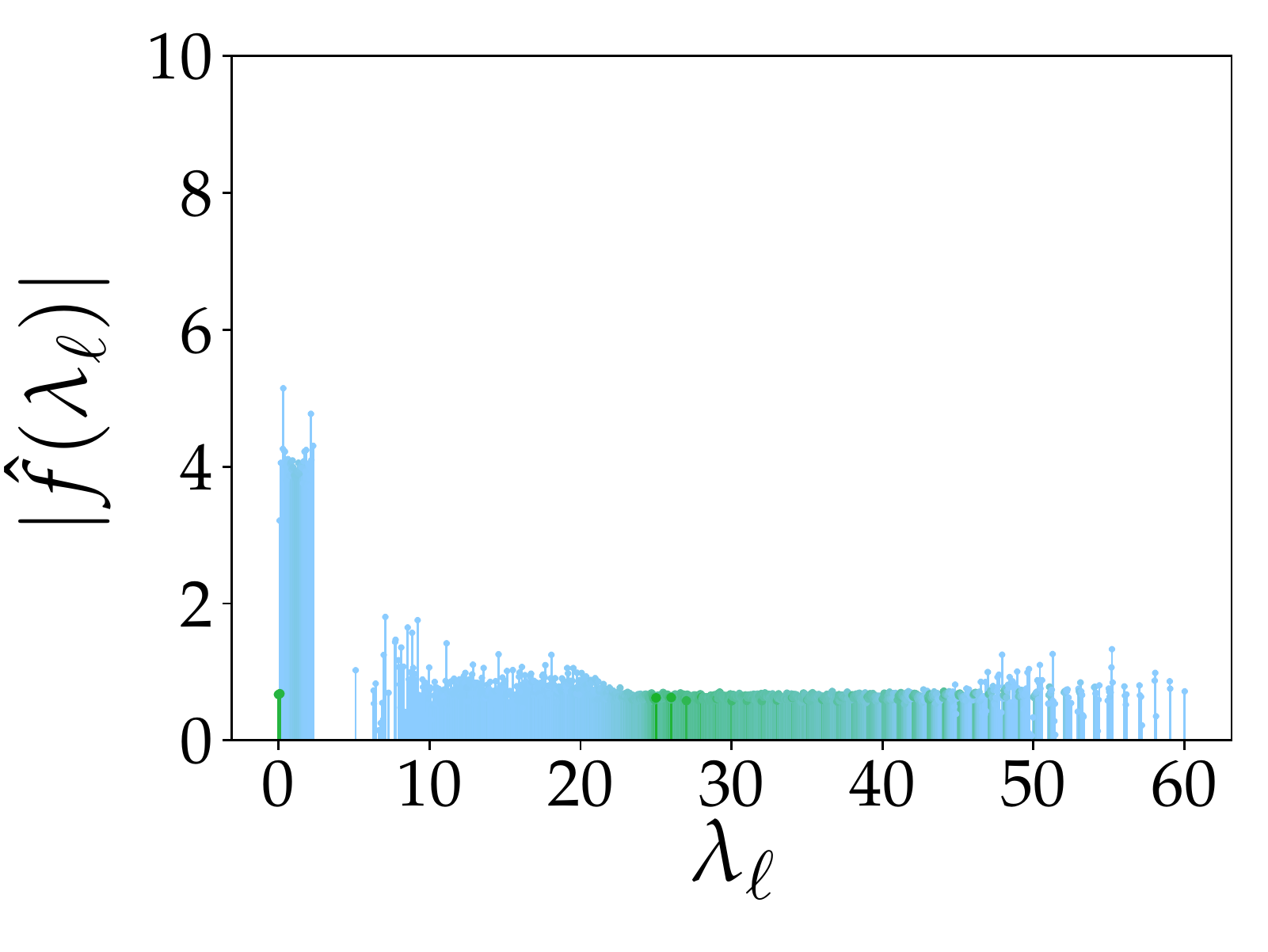}&
            \includegraphics[align=c,width=0.48\linewidth]{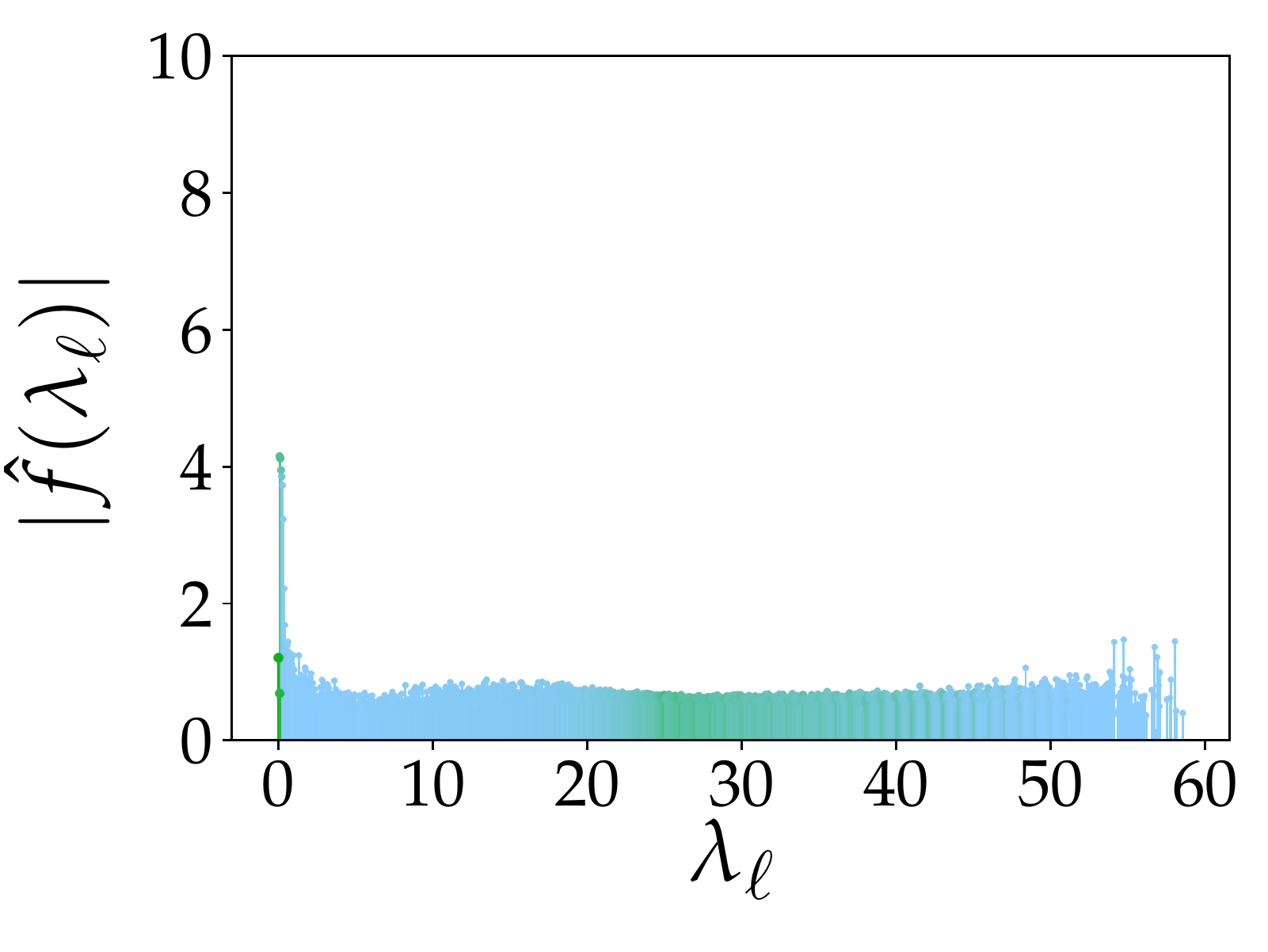}\\
			\includegraphics[align=c,width=0.48\linewidth]{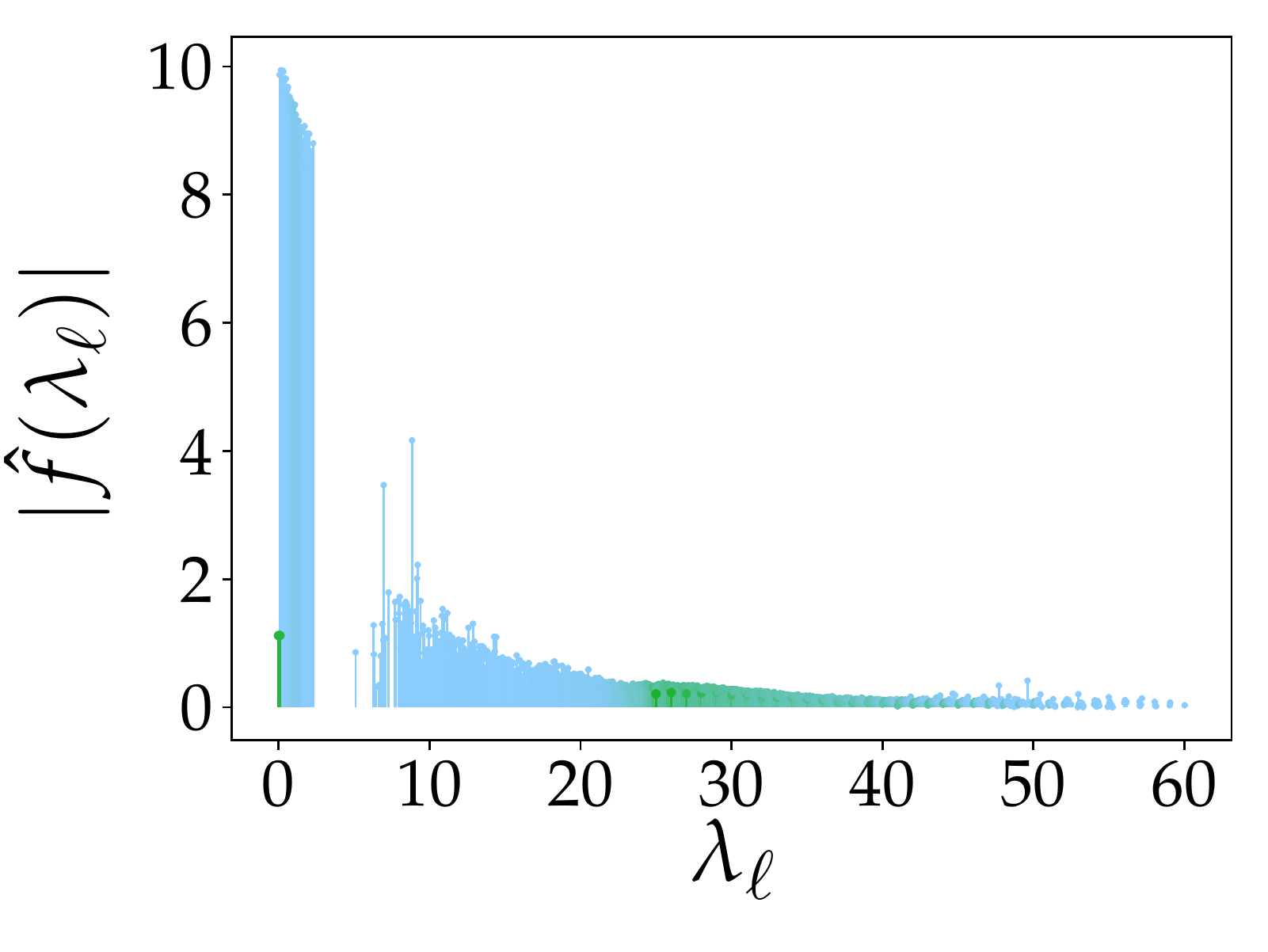}&
			\includegraphics[align=c,width=0.48\linewidth]{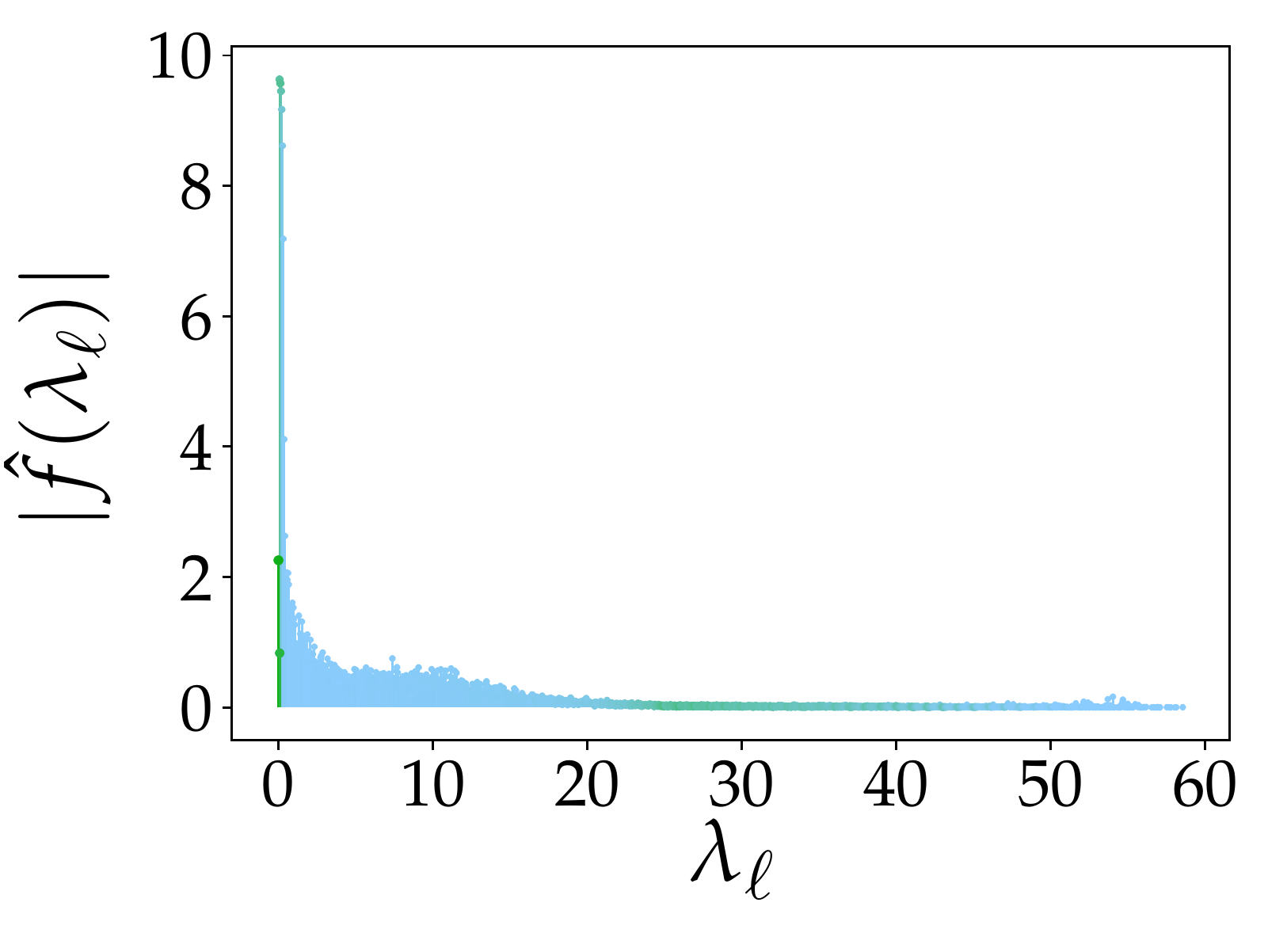}\\
			\includegraphics[align=c,width=0.48\linewidth]{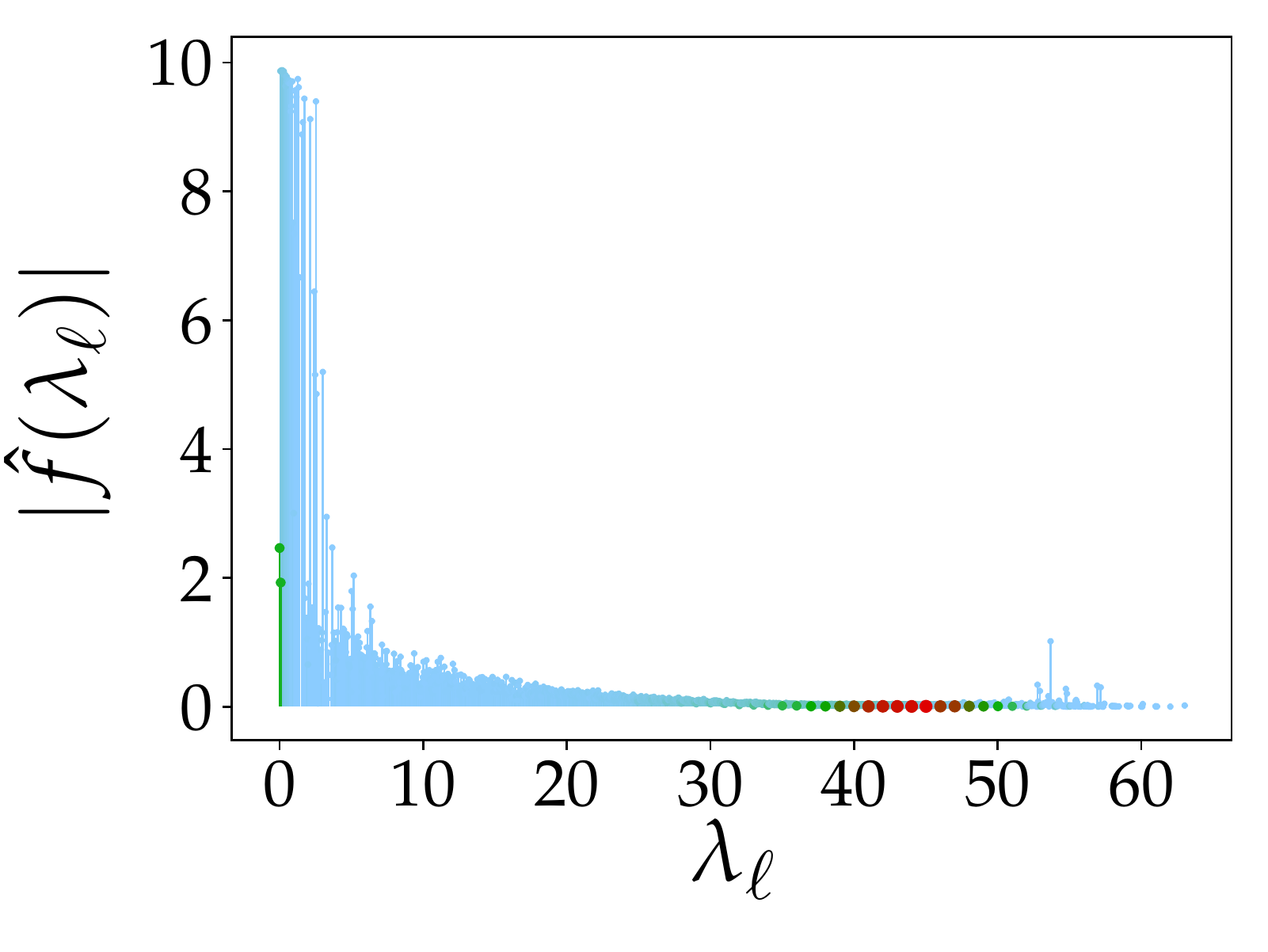}&
			\includegraphics[align=c,width=0.46\linewidth]{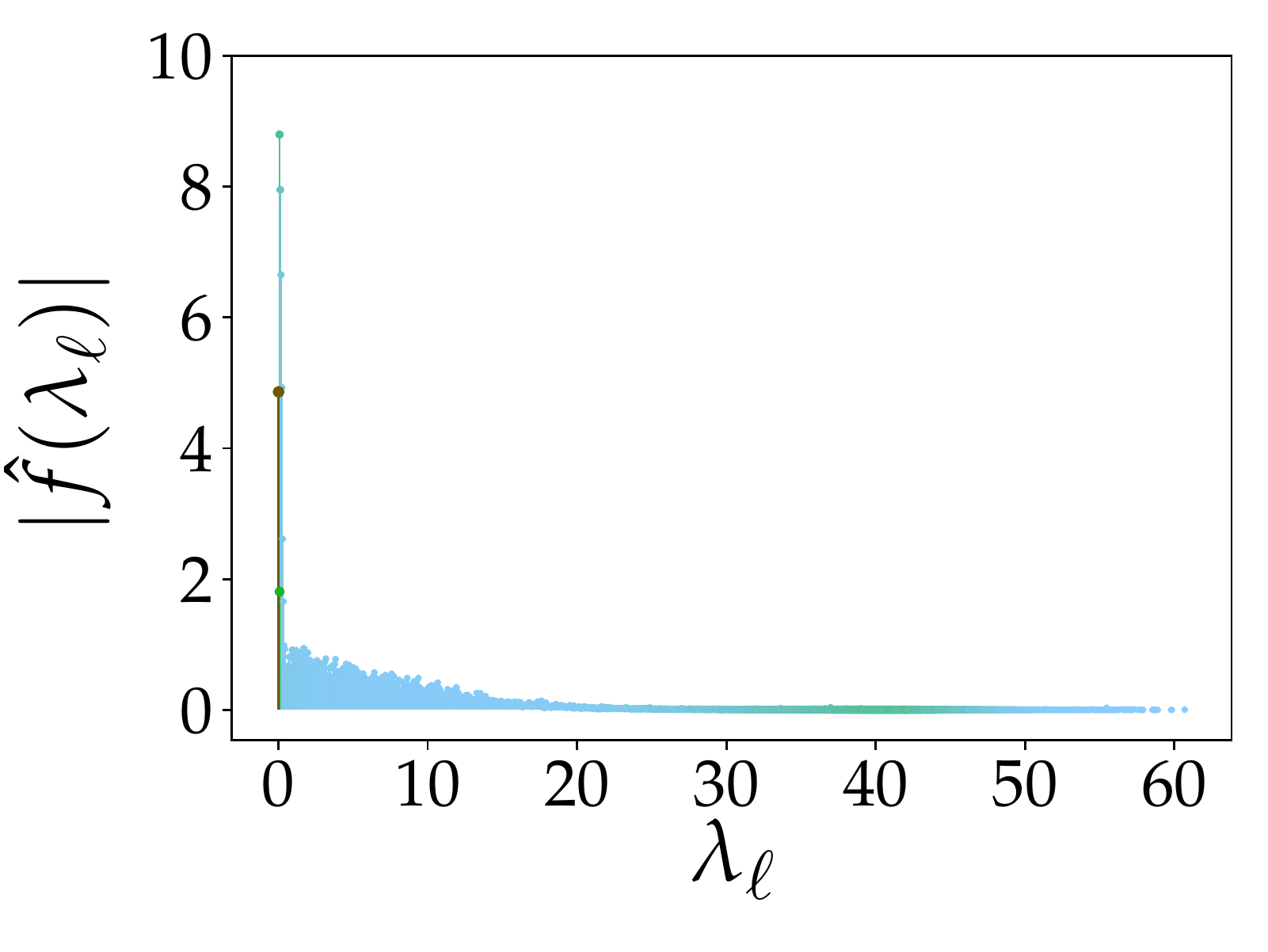}\\
			\multicolumn{2}{c}{\includegraphics[align=c,width=0.7\linewidth]{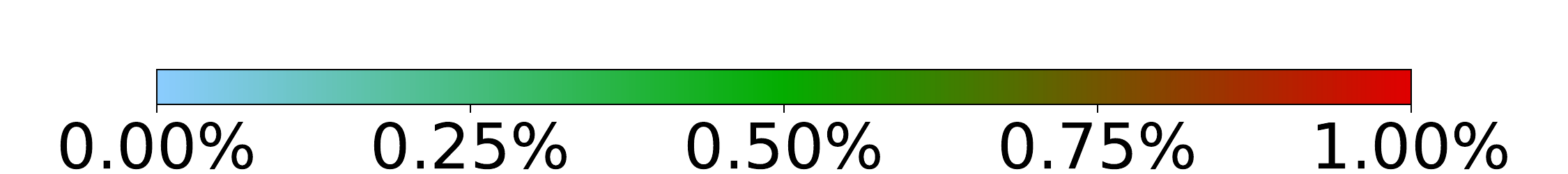}}
		\end{tabu}
		\caption{The magnitude of the Graph Fourier Transform Coefficients for \textit{\textbf{Spambase}} Dataset. The density of each Eigenvalue $\lambda_\ell$ across all experiments on the testing sets is represented through colormaps. Top row shows the result after initialization and before GLR (G-Net output) and the second and third row show the result after the first ($r=1$) and the second iteration ($r=2$), respectively.}
		\label{fig:spambase_gft}
	\end{figure}

	Graph Fourier Transform (GFT) is another approach to represent the smoothness and connectivity of an underlying graph. As in \cite{shuman2013_gsp}, we visualize the magnitude of the GFT coefficients in \Cref{fig:phoneme_gft,fig:spambase_gft} along the graph update iterations.

	In accordance with \cite{shuman2013_gsp}, where it is shown that the magnitude of GFT coefficients decay rapidly for a smooth signal, in our results, we can clearly observe that the magnitude of GFT coefficients is decaying more rapidly along spectral frequencies once the graph is updated (in iteration $r=1$). Furthermore, comparing the visualization results in the first and the second column, we can see that the graph weighting (W-Net) smooths the graph data, with low frequency components becoming more prominent.
	
	\subsubsection{Classification Error Rate Comparison}
	\label{ss:5Class}

	\begin{table}[!htbp]
		\centering
		\caption{Classification Error Rate (\%) For \textit{\textbf{Phoneme}} Dataset}
		\label{tab:phoneme}
		\begin{tabu} to \linewidth{|c|X[c]|X[c]|X[c]|X[c]|X[c]|X[c]|}
			\hline
			\% label noise & 0 & 5 & 10 & 15 & 20 & 25 \\ \hline
			SVM-RBF & 18.33 & 18.75 & 19.13 & 19.57 & 20.07 & 20.87 \\ \hline
			CNN & 17.58	& 17.77	& 18.00	& 18.57	& 19.01	& 20.00 \\ \hline
			DynGraph-CNN & 17.66 & 19.04 & 20.80 & 22.44 & 25.20 & 28.84 \\ \hline
			DML-KNN & 17.04 & 17.54 & 17.82 & 18.58 & 19.64 & 21.00 \\ \hline
			DML-KNN-s & 17.01 & 17.49 & 17.71 & 18.43 & 19.24 & 20.41 \\ \hline
			LN-Robust-SVM-RBF$^{*}$ & 18.57 & 19.42 & 19.65 & 19.70 & 20.03 & 20.28 \\ \hline
			Graph-Hybrid$^{*}$ & 22.01 & 23.77 & 25.58 & 27.97 & 30.33 & 33.39 \\ \hline
			CNN-Savage$^{*}$ & 17.52 & 17.72 & 18.04 & 18.51 & 19.02 & 19.87 \\ \hline
			CNN-BootStrapHard$^{*}$ & 17.46 & 17.72 & 18.00 & 18.31 & 18.84 & 20.15 \\ \hline
			CNN-D2L$^{*}$ & 17.47 & 17.80 & 17.96 & 18.41 & 18.91 & 20.04 \\ \hline
			\textbf{DynGLR-G-2}$^{*}$ & 17.04 & 17.50 & 17.70 & 18.34 & 18.81 & 20.03  \\ \hline
            \textbf{DynGLR-G-12}$^{*}$ & 16.93 & 17.36 & 17.64 & 18.23 & 18.52 & 19.59 \\ \hline
            \textbf{DynGLR-G-12s}$^{*}$ & 16.89 & 17.36 & 17.62 & 18.21 & 18.52 & 19.54 \\ \hline
            \textbf{DynGLR-G-1232}$^{*}$ & 16.90 & 17.29 & 17.36 & 18.16 & 18.48 & 19.47 \\ \hline
			\textbf{DynGLR-G-12312}$^{*}$ & 16.87 & 17.19 & 17.34 & 18.03 & 18.38 & 19.43 \\ \hline
			\textbf{DynGLR-G-12312s}$^{*}$ & \textbf{16.87} & \textbf{17.18} & \textbf{17.32} & \textbf{17.91} & \textbf{18.24} & \textbf{19.18} \\ \hline
		\end{tabu}
	\end{table}
	
	\begin{table}[!htbp]
		\centering
		\caption{Classification Error Rate (\%) For \textit{\textbf{Magic}} Dataset}
		\label{tab:magic}
		\begin{tabu} to \linewidth{|c|X[c]|X[c]|X[c]|X[c]|X[c]|X[c]|}
			\hline
			\% label noise & 0 & 5 & 10 & 15 & 20 & 25 \\ \hline
			SVM-RBF & 18.42 & 19.07 & 19.63 & 20.18 & 20.61 & 21.13 \\ \hline
			CNN & 16.45 & 16.87 & 16.91 & 17.62 & 18.09 & 18.86 \\ \hline
			DynGraph-CNN & 17.74 & 18.69 & 19.33 & 21.05 & 24.15 & 27.40 \\ \hline
			DML-KNN & 15.33 & 15.51 & 15.80 & 15.94 & 16.83 & 18.29 \\ \hline
			DML-KNN-s & 15.33 & 15.51 & 15.78 & 15.89 & 16.58 & 17.08 \\ \hline	
			LN-Robust-SVM-RBF$^{*}$ & 18.57 & 18.70 & 18.80 & 19.05 & 19.39 & 19.82 \\ \hline
			Graph-Hybrid$^{*}$ & 24.82 & 25.92 & 27.23 & 28.84 & 30.79 & 33.23 \\ \hline
			CNN-Savage$^{*}$ & 16.31 & 16.74 & 16.99 & 17.41 & 18.10 & 18.87 \\ \hline
			CNN-BootStrapHard$^{*}$ & 16.34 & 16.89 & 17.02 & 17.46 & 18.13 & 18.65 \\ \hline
			CNN-D2L$^{*}$ & 16.34 & 16.79 & 17.21 & 17.48 & 18.20 & 18.75 \\ \hline
			\textbf{DynGLR-G-2}$^{*}$ & 15.35 & 15.51 & 15.77 & 15.94 & 16.75 & 18.03  \\ \hline
            \textbf{DynGLR-G-12}$^{*}$ & 15.22 & 15.47 & 15.68 & 15.85 & 16.60 & 17.33 \\ \hline
            \textbf{DynGLR-G-12s}$^{*}$ & 15.22 & 15.47 & 15.68 & 15.83 & 16.52 & 16.91 \\ \hline
            \textbf{DynGLR-G-1232}$^{*}$ & 15.22 & 15.46 & 15.66 & 15.85 & 16.58 & 17.18 \\ \hline
			\textbf{DynGLR-G-12312}$^{*}$ & 15.22 & 15.46 & 15.66 &  15.85 & 16.55 &17.17  \\ \hline
			\textbf{DynGLR-G-12312s}$^{*}$ & \textbf{15.22} & \textbf{15.45} & \textbf{15.65} & \textbf{15.83} & \textbf{16.49} & \textbf{16.85} \\ \hline
		\end{tabu}
	\end{table}
	
	\begin{table}[!htbp]
		\centering
		\caption{Classification Error Rate (\%) For \textit{\textbf{Spambase}} Dataset}
		\label{tab:spambase}
		\begin{tabu} to \linewidth{|c|X[c]|X[c]|X[c]|X[c]|X[c]|X[c]|}
			\hline
			\% label noise & 0 & 5 & 10 & 15 & 20 & 25 \\ \hline
			SVM-RBF & 8.09 & 8.50 & 8.98 & 9.75 & 10.68 & 11.49 \\ \hline
			CNN & 7.69 & 8.27 & 8.89 & 9.85 & 10.8 & 12.47 \\ \hline
			DynGraph-CNN & 8.33 & 9.01 & 10.45 & 12.68 & 16.78 & 22.34 \\ \hline
			DML-KNN & 7.84 & 8.41 & 8.49 & 8.98 & 9.83 & 11.02 \\ \hline
			DML-KNN-s & 7.81 & 7.35 & 8.42 & 8.86 & 9.74 & 10.34 \\ \hline
			LN-Robust-SVM-RBF$^{*}$ & 7.84 & 8.38 & 8.89 & 9.68 & 10.56 & 11.32 \\ \hline
			Graph-Hybrid$^{*}$ & 18.34 & 19.19 & 20.37 & 21.85 & 24.17 & 26.68 \\ \hline
			CNN-Savage$^{*}$ & 8.04 & 8.36 & 8.90 & 9.80 & 10.42 & 12.13 \\ \hline
			CNN-BootStrapHard$^{*}$ & 7.69 & 8.30 & 9.24 & 9.68 & 10.05 & 12.01 \\ \hline
			CNN-D2L$^{*}$ & 7.73 & 8.46 & 9.05 & 9.87 & 10.96 & 12.17 \\ \hline	
			\textbf{DynGLR-G-2}$^{*}$ & 7.84 & 8.37 & 8.42 & 8.95 & 9.85 & 10.83 \\ \hline
			\textbf{DynGLR-G-12}$^{*}$ & 7.73 & 8.13 & 8.37 & 8.78 & 9.44 & 9.82 \\ \hline
			\textbf{DynGLR-G-12s}$^{*}$ & 7.72 & 8.11 & 8.35 & 8.75 & 9.35 & 9.63 \\ \hline
			\textbf{DynGLR-G-1232}$^{*}$ & 7.65 & 8.05 & 8.22 & 8.67 & 9.15 & 9.56 \\ \hline
			\textbf{DynGLR-G-12312}$^{*}$ & 7.55 & 7.99 & 8.21 & 8.64 & 9.01 & 9.18  \\ \hline
			\textbf{DynGLR-G-12312s}$^{*}$ & \textbf{7.55} & \textbf{7.94} & \textbf{8.18} & \textbf{8.61} & \textbf{8.96} & \textbf{9.13} \\ \hline
		\end{tabu}
	\end{table}
	\cref{tab:phoneme,tab:magic,tab:spambase} show comparison between the proposed DynGLR networks and all the benchmarks in terms of classification error rate. From the tables, it can be seen that the proposed DynGLR networks outperform all the benchmarks.
	
	\subsection{Summary of findings}
	\label{sec:summary}
	
	
	  The DynGLR-G networks at the bottom of performance tables \cref{tab:phoneme,tab:magic,tab:spambase} show the outcomes of the ablation study. Specifically: 
	  
	  \begin{itemize}
	  \item DynGLR-G-12 consistently outperforms DynGLR-G-2, showing the effect of edge-weighting. 
	  
	  \item The improvement due to graph update can be observed between DynGLR-G-12 and DynGLR-G-1232
	  
	  \item By comparing DynGLR-G-1232 and DynGLR-G12312, we observe small gains, except for low noise in Spambase dataset, due to iterative design, incorporating edge weighting.
	  
	  \item By comparing DynGLR-G-2 and DML-KNN, we can observe performance improvements due to replacing KNN-based classification with GLR; larger gains can be observed as noise level increases.
	  
	  \item Semi-supervised classifiers DML-KNN and all DynGLRs with sampling (DynGLR-G-12s and DynGLR-G-12312s) benefit from rank-sampling, which also reduces the scale of training set without sacrificing the performance.  
	  
	  \end{itemize}
	  
	  Furthermore, our findings are that the importance of the following algorithmic steps, in order of largest to least importance, to the performance can be summarized as: (I) iterative graph update Eq.(\ref{eq:graphupdate}) - disconnect $\mathbf{Q}$-edges and connect/reconnect $\mathbf{P}$-edges based on the restored labels after each GLR iteration to refine the graph structure, (II) edge convolution operation - performing feature and denoised label aggregation on neighboring nodes provides richer and smoother inputs and results in a spatially sparse graph, (III) edge attention function (Eq.~\ref{eq:edgeatt}) - to better reflect node-to-node correlation, regularizing CNN training by weighting the edge loss based on classifier signal changes before and after GLR.
	
	\section{Conclusions}
	In this paper, we introduce an end-to-end iterative graph-based deep learning architecture design to tackle the overfitting problem caused by the effects of noisy training labels. We first propose a CNN-based graph generator G-Net to build an initial graph. Relying on the proposed graph-based regularized loss functions, we then propose a graph-based classifier W-Net to perform online label denoising of the training samples that potentially have noisy labels. Based on the denoised training labels, we update the underlying graph structure by learning the proposed graph update U-Net. Finally, we learn a refined graph-based classifier W-Net to perform classification using the updated underlying graph structure. The validation on three different binary classification datasets demonstrate that our proposed architecture outperforms the state-of-the-art classification methods when partial training labels are incorrect. Furthermore, the rank-sampling method is proven to be another enhancement for this semi-supervised classification problem.
	
	\section{Acknowledgement}
	\begin{tabular}{@{\hskip -8pt}m{0.08\linewidth}@{\hskip 4pt}m{0.9\linewidth}}
		\includegraphics[width=\linewidth]{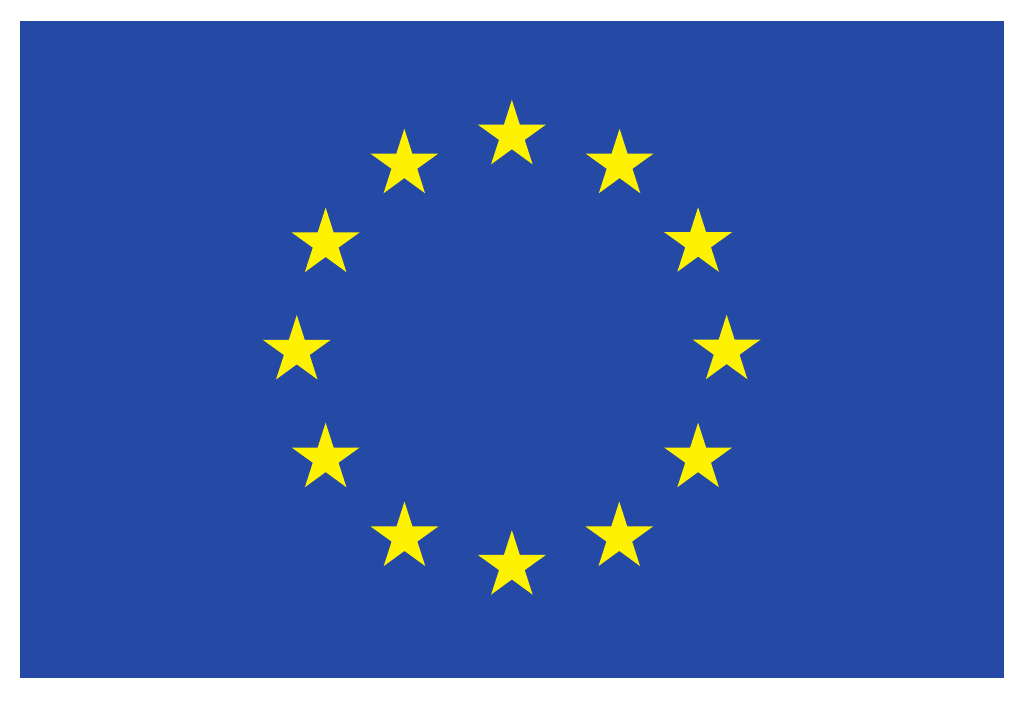}& This project has received funding from the European Union’s Horizon 2020 research and innovation programme under the Marie Skłodowska-Curie grant agreement No 734331. The University of Strathclyde gratefully acknowledges the support of NVIDIA Corporation with the donation of the Titan Xp GPU used for this research.
	\end{tabular}
	

	\printbibliography
\end{document}